\documentclass[lettersize,journal]{IEEEtran}
\IEEEoverridecommandlockouts

\usepackage{amsfonts}
\usepackage{cite}
\usepackage{amsmath}
\usepackage{algorithmic}
\usepackage{graphicx}
\usepackage{textcomp}
\usepackage[dvipsnames,table,xcdraw]{xcolor}
\usepackage{siunitx}
\usepackage{tabularx,booktabs}
\usepackage{multirow}
\usepackage{multicol}
\usepackage{caption}
\usepackage{subcaption}
\usepackage{cleveref}
\usepackage{balance}
\usepackage{lastpage}
\usepackage{tabulary}
\usepackage{breqn}
\usepackage{bbm}
\usepackage{lipsum}
\usepackage{algorithm}
\usepackage{fancyhdr} 
\newlength\myindent
\newcolumntype{C}{>{\centering\arraybackslash}X} 
\newcolumntype{L}{>{\raggedright\arraybackslash}X}

\def\BibTeX{{\rm B\kern-.05em{\sc i\kern-.025em b}\kern-.08em
    T\kern-.1667em\lower.7ex\hbox{E}\kern-.125emX}}
    
\graphicspath{{images/}}

\newcommand{\noteblue}[1]{\textcolor{black}{{ #1}}}

\newcommand{\Hao}[1]{{#1}}

\begin{document}
\title{Smart Home Energy Management:\\ VAE-GAN synthetic dataset generator and Q-learning}

\author{Mina Razghandi,~\IEEEmembership{Graduate Student Member,~IEEE,} Hao Zhou, \IEEEmembership{Graduate Student Member, IEEE}, Melike Erol-Kantarci ~\IEEEmembership{Senior Member,~IEEE,} Damla Turgut, ~\IEEEmembership{Senior Member,~IEEE}
}


\renewcommand\floatpagefraction{.9}
\renewcommand\topfraction{.9}
\renewcommand\bottomfraction{.9}
\renewcommand\textfraction{.1}
\setcounter{totalnumber}{50}
\setcounter{topnumber}{50}
\setcounter{bottomnumber}{50}
\definecolor{cadmiumgreen}{rgb}{0.0, 0.42, 0.24}
\maketitle

\thispagestyle{fancy}            
\chead{This paper has been accepted by IEEE Transaction on Smart Grid. } 

\renewcommand{\headrulewidth}{0pt}      
\pagestyle{plain} 

\begin{sloppypar}
\begin{abstract}
Recent years have noticed an increasing interest among academia and industry towards analyzing the electrical consumption of residential buildings and employing smart home energy management systems (HEMS) to reduce household energy consumption and costs. HEMS have been developed to simulate the statistical and functional properties of actual smart grids. Access to publicly available datasets is a major challenge in this type of research. The potential of artificial HEMS applications will be further enhanced with the development of time series that represent different operating conditions of the synthetic systems.

In this paper, we propose a novel variational auto-encoder-generative adversarial network (VAE-GAN) technique for generating time-series data on energy consumption in smart homes. We also explore how the generative model performs when combined with a Q-learning-based HEMS. We tested the online performance of Q-learning-based HEMS with real-world smart home data. To test the generated dataset, we measure the Kullback–Leibler (KL) divergence, maximum mean discrepancy (MMD), and the Wasserstein distance between the probability distributions of the real and synthetic data. Our experiments show that VAE-GAN-generated synthetic data closely matches the real data distribution. Finally, we show that the generated data allows for the training of a higher-performance Q-learning-based HEMS compared to datasets generated with baseline approaches.
\end{abstract}


\section{Introduction}
\label{Introduction}
\IEEEPARstart{T}{he} design of modern smart homes must take into consideration energy-efficient technologies and renewable energy sources. Recent advances in smart grid research include new technologies and strategies for managing energy generation, storage, supply, and demand~\cite{aliero2021smart}. Smart homes, as a critical component of the smart grid, are projected to increase household energy efficiency, save energy costs, and improve user comfort~\cite{b1}. Utilities and smart home controllers\noteblue{can} acquire information from smart meters for use in forecasting consumption and generation, demand-side management, and economic power dispatch~\cite{b2}. Thus, acquiring fine-grained data about the living environment is becoming an essential precondition for a smart home.

It has become increasingly common to use artificial intelligence and statistical analysis for demand response and energy management tasks requiring accurate short-term (second-to-minute scale) representations of load behavior. These techniques usually require large datasets of representative data for training. However, the collection of such data presents significant security and privacy challenges, with relatively few high-quality publicly available datasets~\cite{zhang2018gan}.

\noteblue{Therefore, synthetic data generation become} an attractive alternative to training machine learning algorithms that can decide, for example, the optimal time for implementing demand response or charging an EV~\cite{z1}. A variety of techniques had been used to generate such datasets, including Markov chains~\cite{b6}, statistical models~\cite{b8}, and physical simulator-based methods~\cite{b7}.
A major drawback of these methods can be attributed to targeting a limited number of problems, applications tailored to specific scenarios, and the lack of scalability. 
\noteblue{Specifically, these methods require case-by-case analyses for each device, user behavior, and environment. But it is impractical to model all the detailed energy consumption behaviors and state transitions when a large number of user devices are involved.}
Recent studies, on the other hand, have utilized deep learning-based approaches that can be applied to large-scale datasets and can be conducted directly on raw data, without the need for further analysis~\cite{b10}.

In our recent paper~\cite{razghandi2022variational}, \noteblue{our approach involved generating electric load profiles and PV generation synthetic data for a smart home using variational autoencoder-generative adversarial networks (VAE-GAN), and we compared the distributions of synthetic data for the VAE-GAN model and vanilla GAN network to the real data distributions. According to the statistical metrics such as KL divergence, Wasserstein distance, and maximum mean discrepancy, the distribution of data points was extremely close between synthetic data and real data for both PV power generation and energy consumption load profiles. Nevertheless, it is a key defining characteristic of a realistic time series to maintain an awareness of temporal differences between synthetic and real-time series. For instance, If PV power production shifted earlier or later in the day rather than following sunrise and sunset patterns, the distribution would still be similar to the real data, but it \noteblue{would not} be realistic. The addition of electric vehicles (EV) charging load consumption synthetic data could provide additional insights and value in understanding smart home electricity consumption patterns. Also, no experiments were conducted to investigate if the synthetic dataset is applicable in real-life practice.}
To address these shortcomings, we propose a\noteblue{revised architecture for the variational autoencoder-generative adversarial network (VAE-GAN) technique for generating time-series data of smart homes from a variety of synthetic data sources and investigate the performance of synthetic data generative models in the presence of control operations. } 
In contrast to the schemes mentioned above, this strategy enables the learning of various types of data distributions in a smart home, including electric load profiles, PV power generation, and EV charging load consumption, and can subsequently generate plausible samples without carrying out any prior analysis before the training phase. 
\noteblue{The VAE-GAN architecture is favored in this model since it allows us to fine-tune the regulated latent space that influences the generated output, as opposed to vanilla GAN, in which the generator maps the input noise to the generated output and makes it susceptible to mode collapse, in which the generator repeatedly produces the same optimal input sequence to mislead the discriminator~\cite{salimans2016improved}.} Furthermore, we compare one model-based and two data-driven generative models, including Gaussian Mixture Model (GMM), vanilla GAN, and VAE-GAN.

Taking advantage of the synthetic data, we propose a Q-learning-based smart home energy management system (HEMS). The generated data is used for the offline training of Q-learning HEMS agents, which aims to maximize long-term management profit. Then, the trained agents are tested online in an environment based on real-world data. Finally, we compare online profits to further investigate how the data generation method will affect the HEMS performance. The idea behind this scheme is that we assume good-quality synthetic data in the offline training can better prepare the agent for real-world operations\cite{b15}. Thus, the performance of the Q-learning-based HEMS is used to demonstrate the quality of the synthetic data.

Compared with conventional model-based optimization algorithms,\noteblue{such as convex optimization,} the reinforcement learning (RL) based method can\noteblue{avoid} the complexity of defining a sophisticated optimization model since the optimization problem can be transformed into a unified Markov decision process (MDP) scheme. On the other hand, most existing HEMS models assume a perfect environment for data collection and algorithm training\cite{z6}. By contrast, the proposed Q-learning HEMS uses synthetic data for training, overcoming the data availability bottleneck.

The main contributions of this paper are as follows:
\begin{itemize}
    \item We propose a VAE-GAN-based scheme to generate high temporal resolution synthetic time-series data for energy consumption in a smart home.
    \item We compare the performance of the proposed approach with techniques based on Gaussian Mixtures and vanilla GANs. 
    \item \Hao{We further investigate the quality of the synthetic data by using it to train a Q-learning-based HEMS model. Evaluating the HEMS agent online in environments based on real-world data, we find that the VAE-GAN method allows higher HEMS profit than other baselines.}
\end{itemize}

The rest of this paper is organized as follows. Section~\ref{RelatedWork} introduces related work. Section ~\ref{ProposedModel} shows the proposed smart home synthetic data generation model and Q-learning-based HEMS model. Section~\ref{Experiments} presents the simulation settings and results, and Section~\ref{Conclusion} concludes the paper.

\section{Related Work}
\label{RelatedWork}
\begin{figure*}[!t]
    \centering
        \centering
        \captionsetup{justification=centering}
        \includegraphics[width=0.97\textwidth]{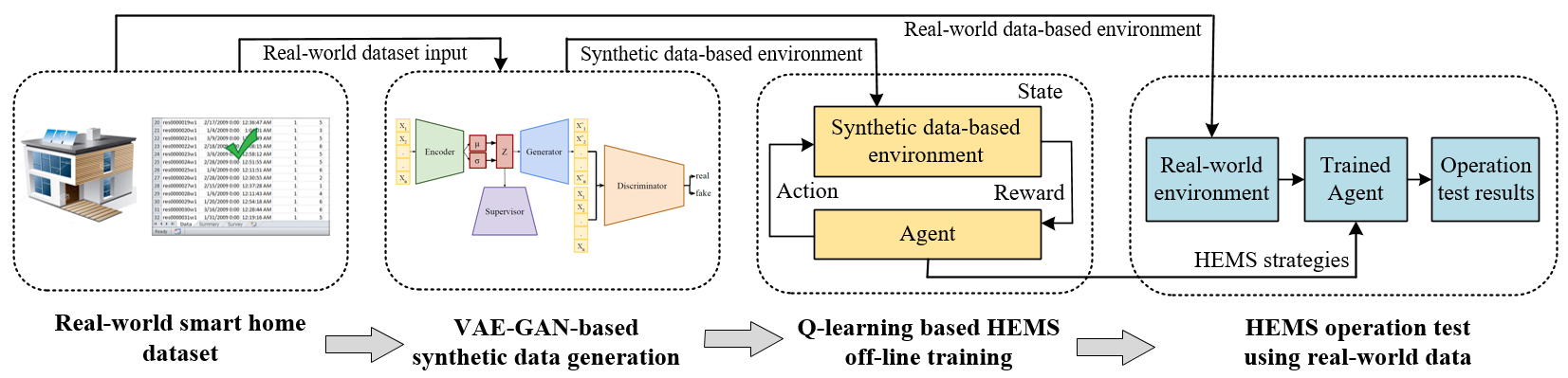}
    \caption[]{\small \Hao{Overall architecture of the proposed scheme}}
        \label{fig-all}
        \vspace{-10pt}
\end{figure*}

The majority of previous work in generating synthetic energy consumption data for smart homes can be classified into model-based or data-driven approaches. Model-based approaches describe the features of household devices with hand-crafted features and mathematical equations. For instance, \cite{b5} proposes a bottom-up residential building energy simulation, which simulates occupant behavior patterns with a Markov chain clustering algorithm. \cite{b6} combined non-intrusive load decomposition and Markov chain methods for user energy consumption simulation. \cite{b7} defines a smart residential load simulator based on MATLAB-Simulink, and it includes dedicated physical models of various household devices. \cite{b8} and \cite{quiros2018statistical} \noteblue{utilize} Gaussian Mixture Modeling to estimate the distribution of the arrival and departure time of electric vehicles (EVs) and to perform temporal modeling of the charging sessions.

On the other hand, recent developments in using GANs have\noteblue{achieved great success} in producing synthetic time-series data. \cite{b11} coupled non-intrusive load decomposition with a conditional GAN to generate synthetic labeled (e.g., appliances) load for a smart home. For smart grid\noteblue{applications}, \cite{b12} proposes a GAN-based approach that captures spatial and temporal correlations between renewable energy sources. \cite{b13} applies a GAN to\noteblue{the} features derived by ARIMA and Fourier transform for generating realistic energy consumption data. \cite{zhang2018gan} uses a GAN to learn the level and pattern features of smart home time-series data, both of which are defined by household consumption and activity, and generate synthesized data with a distribution similar to the real data. \cite{wang2020generating} clusters daily load profiles with known clustering techniques and uses a GAN to synthesize daily loads for each cluster. \cite{buechler2021evgen} proposed a GAN with a Wasserstein loss function to learn temporal and power patterns of EV charging sessions and create a synthetic load dataset.

Reinforcement learning-based energy management models have been extensively investigated in the smart grid. For example, a correlated Q-learning-based approach is proposed in \cite{z1} for microgrid energy management, and a multi-agent reinforcement learning method is introduced in \cite{z3} to minimize the energy cost of different smart home devices. A self-learning HEMS model is presented in \cite{z4}, which includes price forecasting, price clustering, and power alert systems. \cite{z2} introduced a Bayesian deep reinforcement learning method for microgrid energy management under communication failure.

However, most aforementioned make the strong assumption of a perfect data collection and agent training environment. In a real-world environment, fine-grained and good-quality data may be inaccessible due to privacy concerns or measurement errors. As such, synthetic data-based HEMS is much more realistic since the agent can use pre-generated and fine-grained data for training. 



\section{Methodology}
\label{ProposedModel}

\noteblue{The overall architecture of the proposed approach is described in Fig. \ref{fig-all}, and the organization of this work is introduced as followings:
\begin{itemize}
    \item \textbf{Synthetic data generation:} Given a real-world smart home dataset, we first apply the VAE-GAN-based method to produce synthetic data, including the smart home load profiles, PV power generation, and EV charging load consumption, which is introduced in the following Section III-A to D. 
    \item \textbf{Q-learning based HEMS training:} Next, the generated synthetic data is used for the offline training of Q-learning based HEMS. The intelligent HEMS agent will interact with a synthetic data-based environment and produces HEMS strategies to maximize long-term profit, which is included in Section III-E. 
    \item \textbf{Real-world HEMS operation test:} Finally, the trained HEMS agent will be tested in a real-world data-based environment. The test is designed to evaluate the real-world performance of the synthetic data-based HEMS, providing an in-depth evaluation of the synthetic data quality. Specifically, it is expected to demonstrate that our synthetic data can be applied to HEMS training, and the agent can obtain a satisfying real-world performance without touching real-world data.   
\end{itemize}}

\noteblue{Among the well-known deep learning-based generative models, GAN and Variational Autoencoders are two of the best known. 
In this network, the encoder module first encodes the input sequence as a Gaussian distribution over the latent space. Then, the supervisor module trains the encoder module to approximate the next time step. In the generator module, the input sequence is reconstructed from the latent space in an attempt to fool the discriminator. On the other hand, the discriminator module trains the generator to create realistic sequences by identifying fake samples.}

\noteblue{In the following, we first introduce the autoencoder(AE) and variational autoencoder (VAE), then we present generative adversarial network (GAN) and VAE-GAN generative models. Finally, we include the Q-learning-based HEMS control.}

\subsection{Autoencoder (AE)}
\label{AEdescription}
An autoencoder is a neural network architecture that uses an unsupervised learning technique to compress the input dimensions into a compressed knowledge representation, called latent space, and then reverses the compressed knowledge representation into the original input dimensions. This architecture has two essential modules: the encoder and the decoder. The encoder module maps the input sequence ($x$) into the meaningful latent space ($z$), and based on $z$ the decoder module outputs a reconstruction of the original input sequence ($\hat{x}$). The original input is unlabeled, but the decoder is trained to make reconstructions as close as possible to the original input by minimizing the reconstruction error, $\mathcal{L}_{reconstr}$, which is the distance between the original input and the subsequent reconstruction.
\begin{dmath}
    \mathcal{L}_{reconstr} = \lVert \hat{x}-x \rVert^{2}
    \label{eq:Lrecon}
\end{dmath}
\noteblue{where $x$ and $\hat{x}$ are original input and the reconstruction, respectively.}

\subsection{Variational Autoencoder (VAE)}
\label{VAEdescription}
After training the autoencoder model, the decoder module can produce new content given a random latent space. Due to the unregulated latent space, which can lead to severe overfitting, the decoder may not be able to interpolate indefinitely in the absence of data points included in the input sequence. Latent space regularity is reliant on the distribution of the input sequence and the dimension of the latent space, hence it is not always guaranteed. As a solution to this limitation, variational autoencoders include a regularization parameter, such as Kullback–Leibler (KL) divergence, in the learning process to achieve the Gaussian distribution of the latent space and avoid overfitting. In this architecture, the VAE encoder ($E$) encodes the input sequence ($x$) as a distribution over latent space defined by Gaussian distribution first moment, mean, and second moment, standard deviation, and parameters. Ultimately, the encoder is trained to minimize the $\mathcal{L}_{prior}$ loss and output a latent space with a Gaussian distribution. In the training process, the VAE minimizes a loss function ($\mathcal{L}_{VAE}$) that consists of two terms: the reconstruction term and the regularization term, which is the KL divergence between the latent space distribution and standard Gaussian distribution.
\begin{dmath}
    \mathcal{L}_{prior} = D_{KL}(E(x) \vert\vert \mathcal{N}(0,1))
    \label{eq:Lprior}
\end{dmath}
\begin{dmath}
    \mathcal{L}_{VAE} = \mathcal{L}_{prior} + \mathcal{L}_{reconstr}
    \label{eq:LVAE}
\end{dmath}
\noteblue{where $D_{KL}$ is the KL divergence, $E$ is VAE encoder, $\mathcal{N}(0,1)$ is the Gaussian distribution, and $\mathcal{L}_{prior}$ is the loss. }

\subsection{Generative Adversarial Network (GAN)}
\label{GANdescription}
Generative models, such as GAN, are used to generate new data samples. This architecture consists of two neural networks: one attempts to detect and learn patterns in the input data and produce a new sample, while the other aims to distinguish between real input data and the synthesized sample. These two networks play against each other and actively seek equilibrium.
\begin{itemize}
    \item \textbf{Generator ($G$):} Takes an input noise $z$, which usually has a normal distribution, and maps that into data samples $G(z;\theta_{g})$, with $\theta_{g}$ as the network parameter.
    \item \textbf{Discriminator ($D$):} $D(x;\theta_{d})$ returns the likelihood of $x$ being classified as a member of the original data, where 0 represents fake data and 1 represents original data. 
\end{itemize}
$G$ and $D$ engage in an adversarial game, equation (\ref{eq:minmax}), where $G$ attempts to fool the $D$ and maximize the final classification error, while $D$ is trained to classify fake data more accurately and minimize the error.
\begin{dmath}
    \underset{G}{min}~\underset{D}{max}   \mathcal{L}_{GAN}(D,G) = E_{x}[log(D(x))] + E_{z}[1-log(D(G(z)))]
    \label{eq:minmax}
\end{dmath}
\noteblue{where $G$ and $D$ has been defined as generator and discriminator, $x$ and $z$ are input sequence and noise, respectively. }

\begin{figure}
    \centering
        \includegraphics[width=0.48\textwidth]{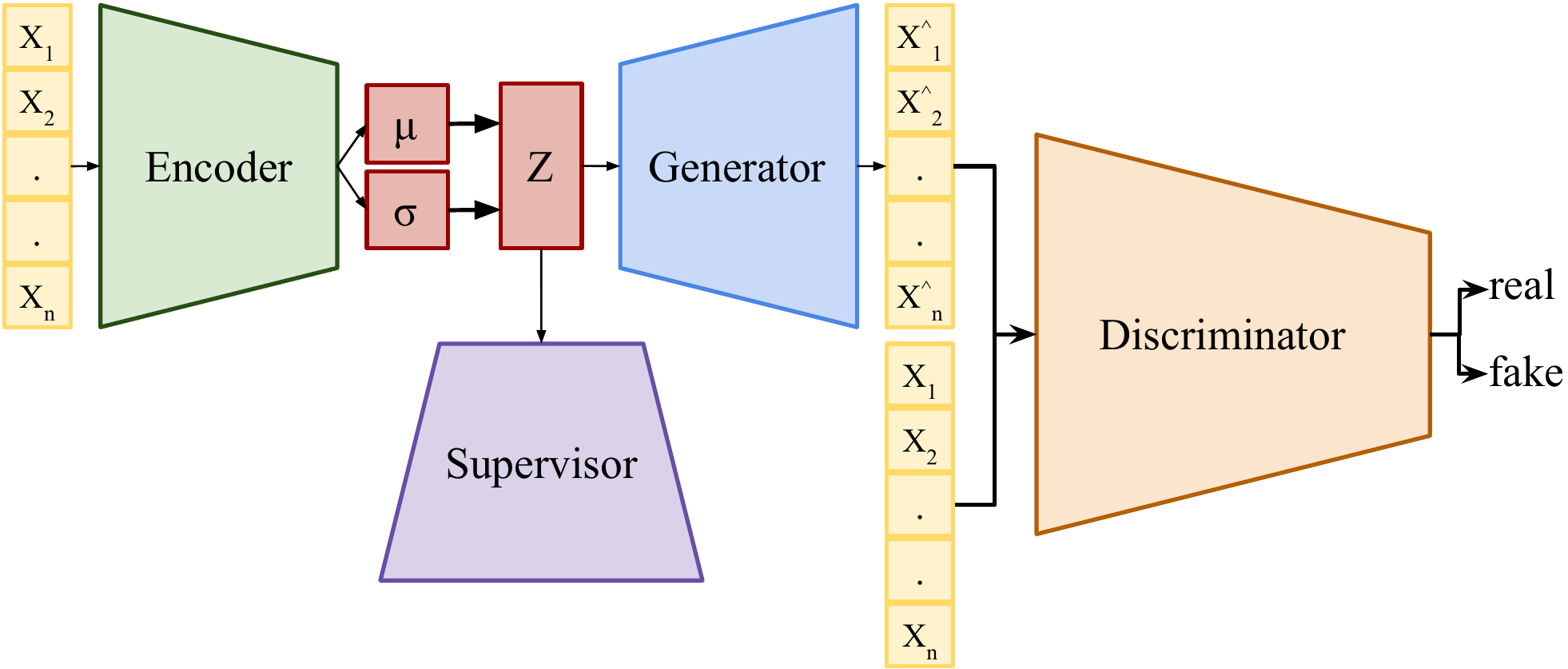}
        \caption{\small {VAE-GAN model architecture. In this network, the encoder module encodes the input sequence as a Gaussian distribution over the latent space, defined by mean and variance vectors. The supervisor module trains the encoder module to approximate the next time step closely in the latent space. In the generator module, the input sequence is reconstructed from the latent space in an attempt to fool the discriminator so the generated sequence is considered real. The discriminator module trains the generator module to create realistic sequences by identifying fake samples from real ones.}}
        \label{fig:model}
\vspace{-5mm}
\end{figure}

\begin{figure}
    \centering
    \begin{minipage}{0.45\linewidth}
        \centering
        \includegraphics[width=\textwidth]{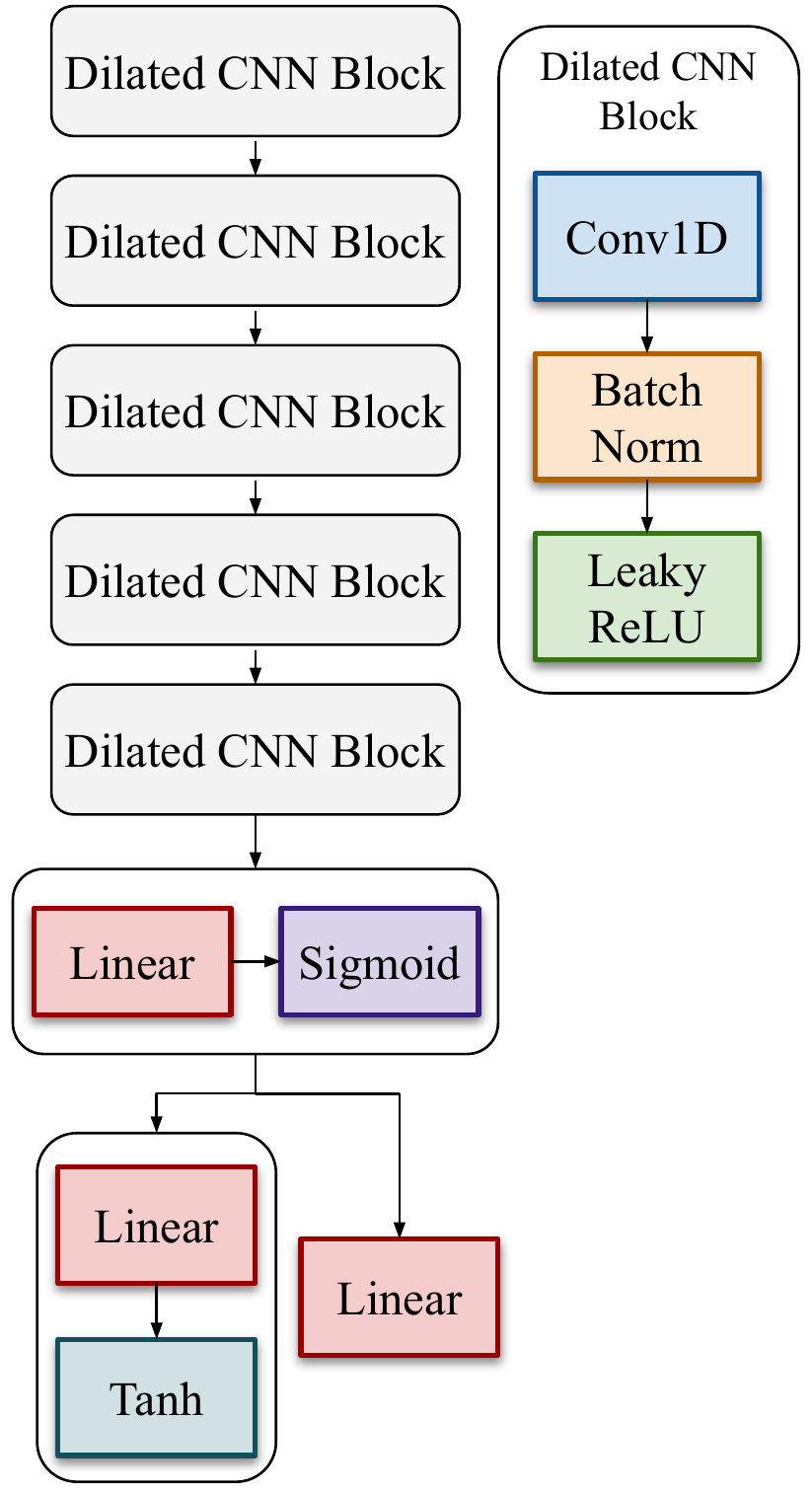} 
        \caption{\small VAE-GAN \textbf{encoder} module structure}
        \label{fig:VAE-enc}
    \end{minipage}\hfill
    \begin{minipage}{0.45\linewidth}
        \centering
        \includegraphics[width=\textwidth]{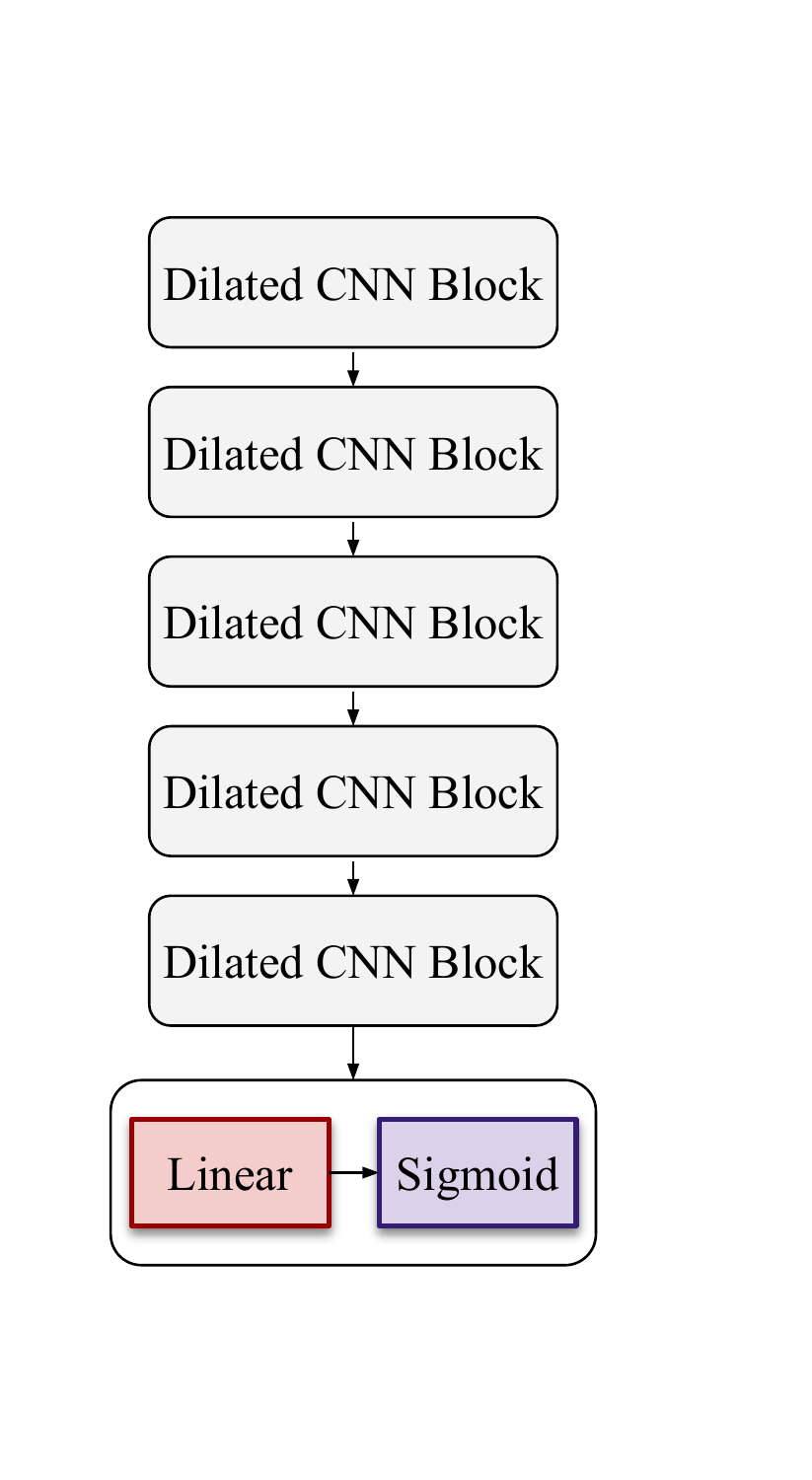} 
        \caption{\small VAE-GAN \textbf{generator}, \textbf{supervisor},  and \textbf{discriminator} modules structure.}
        \label{fig:VAE-gen}
    \end{minipage}
\vspace{-5mm}
\end{figure}

\subsection{Data-driven Generative Model}
\label{VAEGANdescription}
To generate synthetic smart home data we adopted a VAE-GAN architecture, as presented in Fig.~\ref{fig:model} that avoids the shortcomings of the vanilla GAN or VAE. For example, vanilla GAN is prone to mode collapse, a situation where the generator finds an input sequence that fools the discriminator and repeatedly produces it over and over again. The VAE-GAN architecture, introduced by Larsen et al.~\cite{pmlr-v48-larsen16}, generates new data samples based on a regulated latent space rather than producing new data samples from a noise input. The discriminator will further classify the generated sample. Considering that the data for smart homes is time-series data, it is essential that the encoder be able to not only detect the underlying pattern and distribution in the input data but also retain the realistic order in which events occur. As a result, we took TimeGAN's~\cite{yoon2019time} concept of incorporating a supervisor module into our VAE-GAN architecture. We will discuss each module in more detail:   

\noteblue{As opposed to mapping the input sequence into a meaningful latent space}, The \textbf{encoder} module ($E$) here translates the original input sequence $x$ into two vectors that represent the mean ($\mu$) and variance ($\sigma$) of a standard Gaussian distribution. Minimizing $\mathcal{L}_{prior}$ (equation (\ref{eq:Lprior})) forces the encoder to compress the data over a standard Gaussian distribution. $\mu$ and $\sigma$ are mapped into the latent space $z$ using the \textit{reparameterization} technique. The encoder uses five layers of \textbf{Dilated One Dimensional Convolutions} (dilated Conv1D) stacked one on top of the other (see Fig.~\ref{fig:VAE-enc}). This architecture promises similar performance but lower complexity and faster convergence compared to the Long Short-Term Memory (LSTM) networks typically used for time-series analysis. The Conv1D layer architecture resembles the WaveNet architecture~\cite{oord2016wavenet}, as each layer has a stride length of one and kernel size of two, but the dilation rate varies in each layer.
As the second layer dilates at any second of the input sequence (skipping every other timestep), the third layer dilates at any fourth, etc., the lower layers are more focused on short-term dependencies while the higher layers capture long-term dependencies.

The \textbf{supervisor} module computes the distance between the latent representation at the current time step ($z$) and the next time step ($\hat{z}$), minimizing $\mathcal{L}_{supervisor}$. Added to the encoder loss function, equation (\ref{eq:LE}), allows the encoder to approximate the next time step in the latent space, retaining the original sequence of events. 
\begin{dmath}
    \mathcal{L}_{supervisor} = \lVert z-\hat{z} \rVert^{2} 
    \label{eq:Lsupervisor}
\end{dmath}
\begin{dmath}
    \mathcal{L}_{E} = \mathcal{L}_{prior} + \mathcal{L}_{supervisor}
    \label{eq:LE}
\end{dmath}

The \textbf{generator} module ($G$) is trained to reconstruct the original input sequence, given latent space $z$ as the input by minimizing the Mean Squared Error (MSE) between the reconstructed sequence and original input sequence ($\mathcal{L}_{reconstr}+\mathcal{L}_{prior}$). The generator loss also contains the term  $\mathcal{L}_{dG}$,  computed as in equation~(\ref{eq:Lgen}), quantifying the likelihood of the discriminator classifying the reconstructed sequence as a fake sequence. The goal is to make the reconstructed sequence so realistic that it can mislead the discriminator.

\noteblue{
\begin{dmath}
    \mathcal{L}_{dG} = E_{x}[log(D(G(z)))] 
    \label{eq:Ldgen}
\end{dmath}
\begin{dmath}
    \mathcal{L}_{generator} = \mathcal{L}_{prior} + \mathcal{L}_{reconstr} + \mathcal{L}_{dG} + \mathcal{L}_{supervisor}
    \label{eq:Lgen}
\end{dmath}
\begin{dmath}
    \mathcal{L}_{real} = E_{x}[log(D(x))]
    \label{eq:Lreal}
\end{dmath}
\begin{dmath}
    \mathcal{L}_{fake} = E_{z}[1-log(D(G(z)))]
    \label{eq:Lfake}
\end{dmath}
\begin{dmath}
    \mathcal{L}_{noise} = E_{z}[1-log(D(\mathcal{N}(0,1)))
    \label{eq:Lnoise}
\end{dmath}
\begin{dmath}
    \mathcal{L}_{D} = \mathcal{L}_{real} + \mathcal{L}_{fake} + \mathcal{L}_{noise} 
    \label{eq:LD}
\end{dmath}}

The \textbf{discriminator} module ($D$) is responsible for classifying the original input sequence and fake data samples. $D$ minimizes $\mathcal{L}_D$, equation (\ref{eq:LD}), in the training process, which has three terms. $\mathcal{L}_{real}$, equation (\ref{eq:Lreal}), is the likelihood of original input data being classified as fake data, $\mathcal{L}_{fake}$, equation (\ref{eq:Lfake}), is the likelihood of reconstructed data sample being classified as real, and $\mathcal{L}_{{noise}}$, equation (\ref{eq:Lnoise}), is the likelihood of classifying a random noise input as a real data which is added to $\mathcal{L}_D$ to improve the convergence of the discriminator.

\subsection{Q-learning based HEMS control}

\begin{figure}
    \centering
        \centering
        \captionsetup{justification=centering}
        \includegraphics[width=0.45\textwidth]{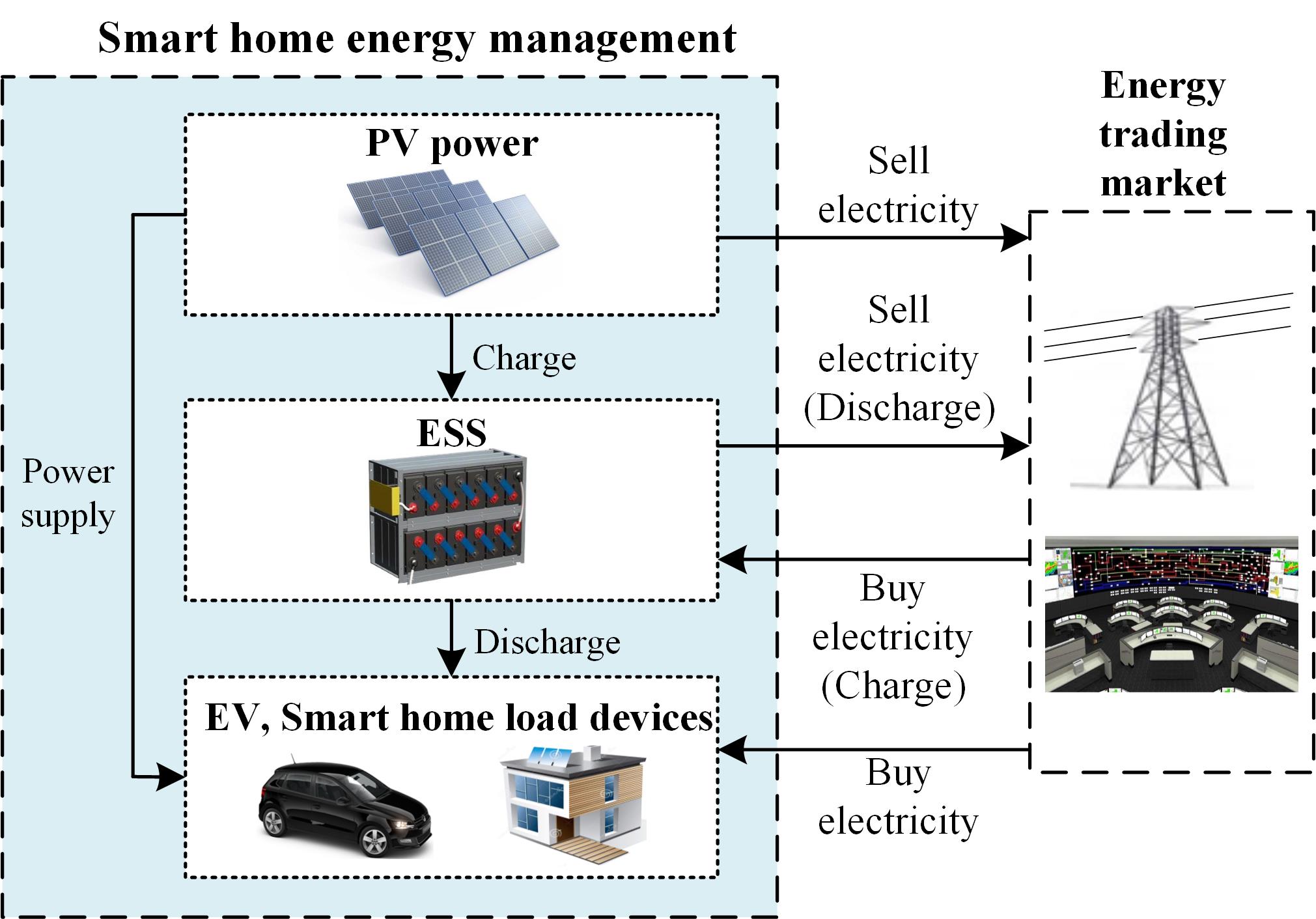}
        \caption{\small The proposed smart home energy management system}
        \label{fig4}
\vspace{-3mm}
\end{figure}

In this section, we introduce the Q-learning-based HEMS model. As shown in Fig.~\ref{fig4}, the components of the smart home involved in the energy production and consumption include the PV panels, loads, EV, and the ESS:
\begin{itemize}
    \item \textbf{PV}: The PV is considered a pure energy supplier. The PV power will first serve the energy demand of EVs and other smart home devices, then the surplus energy can be used for ESS charging or selling to the energy trading market for profit.
    \item \textbf{EV and other smart home loads}: EV and other smart home loads are pure energy consumers. They will first receive the power supply from PV or ESS for operation or buy electricity from the market if the internal power supply is insufficient. 
    \item \textbf{ESS}: Finally, the ESS can be a power supplier when discharging, or a consumer when charging. For charging, it uses the surplus PV power or buys electricity from the market. For discharging, it supplies energy to EVs and smart home devices or sells electricity to the energy trading market for profit.
\end{itemize}

The HEMS model takes advantage of the flexibility of ESS to minimize the total energy cost and maximize profit. The optimization task of the centralized HEMS model is described as follows:
\begin{subequations}\label{e2:main}
\begin{align}
& \text{max} && \sum\nolimits_{t=1}^{T}P^{total}(\gamma p^{sell}_{t}+(1-\gamma)p^{buy}_{t})& \tag{\ref{e2:main}}\\
&  \text{s.t.}           && P^{total}=P_{t}^{ESS}+P_{t}^{PV}-P_{t}^{L}-P_{t}^{EV} & \label{e2:a} \\
&  &&\gamma= \mathbbm{1}\{P^{total}>0\} & \label{e2:d}  \\
&             && P_{t}^{ESS}=P_{ch}q_{t} & \label{e2:c}\\
&             && q_{t}=\left\{
\begin{array}{rcl}
-1     &      &  ESS\quad charges\\
 0     &      &  ESS \quad unchanged\\
 1     &      &  ESS \quad discharges
\end{array} \right. & \label{e2:e}\\
&             && Soc_{t+1}= Soc_{t}-\frac{P_{t}^{ESS}}{C^{ESS}} & \label{e2:f}\\
&             && Soc_{min}\leq Soc_{t} \leq Soc_{max} & \label{e2:g}
\end{align}
\end{subequations}
where $T$ is the total optimization period, $p^{sell}_{t}$ and $p^{buy}_{t}$ represent the price of selling energy to the energy trading market and buying energy from the market, respectively. $P_{t}^{ESS}$, $P_{t}^{PV}$, $P_{t}^{L}$ and $P_{t}^{EV}$ represent the power of ESS, PV, smart home load, and EV charging load at time slot $t$, respectively. $P^{total}$ denotes the total power consumption/demand of the smart home, and (\ref{e2:a}) is the energy balance constraint. $\mathbbm{1}{}$ is an indicator function. $\mathbbm{1}\{P^{total}>0\}=1$ when $P^{total}>0$, which means the agent will sell surplus energy for profit; \noteblue{otherwise}, $\mathbbm{1}\{P^{total}>0\}=0$ if $P^{total}<0$, which means buying energy from the market. $P_{ch}$ is the fixed ESS charging power. $q_{t} = -1,0,1$ when ESS charges, remain unchanged and discharges, respectively. $C^{ESS}$ is the fixed capacity of ESS, and $Soc$ is the ESS state of charge (SOC). Equations (\ref{e2:c}) to (\ref{e2:f}) are ESS operation constraints, and (\ref{e2:g}) is the SOC upper and lower bound constraint.

To transform this optimization task into the context of Q-learning, we define a Markov decision process (MDP) as follows.
\begin{itemize}
    \item \textbf{State:} The agent state is defined as\\ $s_{t}=\{Soc_{t},P_{t}^{PV},P_{t}^{Load}\}$, where $P_{t}^{Load}=P_{t}^{L}+P_{t}^{EV}$ represents the total energy consumption of smart home. 
    \item \textbf{Action:} Based on the total energy demand and PV power generation, the agent decides on the action $a_{t}={q_{t}}$, which indicates the charging, discharging, or remaining unchanged status of ESS.  
    \item \textbf{Reward:} By selecting actions intelligently, the agent intends to maximize the total profit in the optimization period, and the reward function is defined by:
    \begin{equation}
     r_{t}=P^{total}(\gamma p^{sell}_{t}+(1-\gamma)p^{buy}_{t}),   
    \end{equation}
    which is the objective function of our problem formulation equation (\ref{e2:main}).
\end{itemize}

\noteblue{In Q-learning, the agent aims to maximize the long-term expected reward 
\begin{equation} \label{eq-ql}
V^{\pi}(s) =E_{\pi}[\sum_{n=0}^{\infty}\gamma^{n} r(s_{n},a_{n})|s=s_{0}],
\end{equation}
where $s_0$ is the initial state, $n$ is the number of iterations, $E_{\pi}$ is the expected value under action selection policy $\pi$, $r(s_{n},a_{n})$ is the reward of selecting action $a_{n}$ under state $s_{n}$, and $\gamma$ is the reward discount factor. A lower $\gamma$ means focusing on immediate rewards, and a high $\gamma$ indicates that future reward is more important. }

\noteblue{Then, we define the state-action value 
\begin{equation} \label{eq-qupdate}
\begin{aligned}
Q^{new}(s_{n},a_{n})&= Q^{old}(s_{n},a_{n})+\\ \alpha(r_{n}&+\gamma \max\limits_{a} Q(s_{n+1},a) -Q^{old}(s_{n},a_{n})),
\end{aligned}
\end{equation}
where $Q^{old}(s_{n},a_{n})$ and $Q^{new}(s_{n},a_{n})$ are old and new Q-values, respectively, indicating the accumulated reward of selecting action $a_n$ under state $s_n$. $s_{n+1}$ is the state of $n+1$ iteration, and $\alpha$ is the learning rate. A high learning rate will lead to a fast learning process, but the results can be unstable; otherwise, a lower learning rate may result in very slow convergence and a long training time.}

\noteblue{In addition, we use the $\epsilon$-greedy policy for the action selection.
\begin{equation} \label{eq-action}
\pi(s) =\left\{
\begin{array}{ccl}
 \arg~\max\limits_a \, Q(s_{n+1},a) ,  &  rand>\epsilon,\\
  \text{random~action~selection}, &   rand\leq\epsilon.\\
\end{array} \right.
\end{equation}
where $rand$ represents a random number $(0 \leq rand \leq 1)$, and $\epsilon <1$. 
When $rand>\epsilon$, the agent takes greedy policy to select the action with maximum Q-value; otherwise, the action is selected randomly for exploration. $\epsilon$-greedy policy can balance the exploration and exploitation of the Q-learning agent to maximize the long-term expected reward.}

Finally, the Q-learning-based HEMS algorithm is summarized in Algorithm~\ref{Q-LearningAlgo}, which consists of two phases. In the offline training phase, given the generated synthetic data, the intelligent agent is trained to maximize the total profit. Then, in the online test phase, we apply real-world data for online operation, which aims to test the performance of agents that are trained on synthetic data from various datasets.

\begin{algorithm}[!t]
	\caption{Q-learning for HEMS}
	\begin{algorithmic}[1]
		\STATE \textbf{Initialize:} Q-learning and smart home parameters
		\STATE {\textbf{Phase 1: \noteblue{Offline} training using synthetic data:}}
		\begin{ALC@g}
		\STATE {\textbf{Input:} generated synthetic data of smart home load, EV load, and PV power generation.}
		\FOR{$episode=1$ to $E$}
		\STATE{With probability $\epsilon$ choose action $a$ randomly. Otherwise, $a=arg\;max (Q(s,a))$} 
		\STATE {Agent calculates reward based on equation (\ref{e2:main})}. 
		\STATE {Update agent state $\{t,S_{t},w_{g,t}\}$ and Q-value:} 
		\STATE {$Q(s,a)=(1-\alpha)Q(s,a)+\alpha(r+\gamma max Q(s',a'))$}
		\ENDFOR
		\end{ALC@g}
		\STATE {\textbf{Output:} Trained Q-learning based HEMS strategy.}
	    \STATE {\textbf{Phase 2: Online test using real-world data:}}
	    \begin{ALC@g}
		\STATE {\textbf{Input:} Real-world data.}
		\STATE {Deploy the pre-trained Q-learning agent for an operation test.}
		\end{ALC@g}
		\STATE {\textbf{Output:} HEMS profit under real-world data.}
	\end{algorithmic}
	\label{Q-LearningAlgo}
\end{algorithm}

\section{Evaluation Study}
\label{Experiments}

\subsection{Dataset}
\label{Dataset}
\subsubsection{SmartHome Dataset}
The iHomeLab RAPT dataset~\cite{huber2020residential} is a real-world dataset for residential power traces. Five households in Switzerland have been surveyed for 1.5 to 3.5 years with a sampling frequency of 5 minutes. The dataset contains residential electricity consumption and PV energy production, including appliance-level and aggregated household consumption data. Studies of PV power generation show recurring patterns with corresponding noises originating from rain and clouds. Consumption patterns show weekly trends, but with occasional irregular spikes.

The experiment was conducted on residential house D from the dataset. Data cleansing was done by choosing days with full 24-hour records, resulting in 594 days in total. As part of the training process, electrical load and PV power production data were downsampled to a resolution of 15 minutes. \noteblue{For the train and test datasets, a split ratio of 80:20 was adopted. }

\subsubsection{Residential Electric Vehicle Charging Dataset}
Sorensen et al.~\cite{sorensen2021residential} presented a residential electric vehicle charging dataset recorded from apartment buildings. From December 2018 to January 2020, 97 users in Norway reported real-world EV charging experiences. Each charging session includes plug-in time, plug-out time, and charged energy. This dataset provides a synthetic charging load for level-1 charging and level-2 charging assuming 3.6 kW or 7.2 kW of charging power. In our experiments, we interpolate the data for a 15-minute resolution. \noteblue{We note that a higher time scale resolution such as 5-minute will increase the simulation running time, while a lower resolution such as 20 minutes can not capture the environment dynamics. In addition, the proposed synthetic data generator scheme can adapt to any time resolution without loss of generality. }

Our analysis revealed that the user with the largest number of available records has only 62 days of 24-hour data. This data contains only 11 charging sessions, which is insufficient for the training of a deep learning model. To address this issue we generated a larger synthetic dataset by combining the charging data from several users and assuming that the data belonged to an indoor charging station shared by the building's residents, \noteblue{thus accumulating 263 days of historical data, We are thereby able to have more significant charging events for the model, resulting in a more efficient training process}.

\subsection{Baseline Models and Performance Metrics}
\label{baselinemodels}

In our experiments, we compared the performance of our VAE-GAN-based approach with two other state-of-the-art approaches:

\begin{itemize}
    \item \textbf{Gaussian Mixture Model (GMM)}: is a probabilistic approach that partitions data into groups using soft clustering based on multiple multidimensional Gaussian probability distributions. The mean and variance of the distributions are calculated using Expectation-Maximization. Upon fitting the GMM to some data, a generative probabilistic model can sample synthetic data, following the same distribution.
    \item \textbf{Vanilla GAN}: As described in Section~\ref{GANdescription}. 
    \item \textbf{VAE-GAN}: As described in Section~\ref{VAEGANdescription}
\end{itemize}

In the following, we introduce the metrics that we will employ to evaluate the quality of the generated synthetic data.

\subsubsection{Kullback–Leibler (KL) divergence}
\label{kldivergence} 
The Kullback-Leibler divergence (KLD) is one of the most often used measures for evaluating the similarity of two probability distributions. Equation (\ref{eq:kldiv}) is the formal definition of KL divergence, where $x$ and $y$ are sampled data points, and $p(x)$ and $p(y)$ are their respective probability distributions. The KL divergence ranges from $0$ when two probability distributions almost match everywhere, to $\infty$ for completely different distributions.
\begin{equation}
    D_{KL}(p \lVert q) =\sum_{i=1}^{N}p(x_{i})log\left(\frac{p(x_{i})}{q(y_{i})}\right)
    \label{eq:kldiv}
\end{equation}

\subsubsection{Maximum Mean Discrepancy (MMD)}
\label{MMD} 
The MMD approach represents the distance between distributions by the distance between the mean embeddings of features into a reproducing kernel Hilbert space.
Given the distributions $p(x)$ for $\{x_i\}_{i=0}^{N}$ and $q(y)$ for $\{y_j\}_{j=0}^{M}$, MMD is calculated as follows:
\begin{dmath}
    MMD(p,q)^2 = \frac{1}{N^2} \sum_{i=1}^{N} \sum_{j=1}^{N} K(x_{i},x_{j})- \frac{2}{MN} \sum_{i=1}^{N} \sum_{j=1}^{M} K(x_{i},y_{j})+ \frac{1}{M^2} \sum_{i=1}^{M} \sum_{j=1}^{M} K(y_{i},y_{j})
    \label{eq:mmd}
\end{dmath}
\begin{equation}
    K(x,y) = \exp\left(\frac{-{\lVert x-y \rVert}^2}{2\sigma^2}\right)
    \label{eq:rbf}
\end{equation}

\subsubsection{Wasserstein Distance}
\label{Wdistance} 
Intuitively, the Wasserstein distance,\noteblue{also called the earth mover's distance,} models a probability distribution as a pile of soil, and computes how much soil needs to be moved to transform one probability distribution to the other.
\begin{dmath}
    l_{1}(p,q) = \underset{\pi \in \Gamma(p,q)}{inf} \int_{\mathbb{R} \times \mathbb{R}} \vert x-y \vert d \pi(x,y)
    \label{eq:wass}
\end{dmath}

Equation (\ref{eq:wass}) represents the formal definition of Wasserstein distance, where $p(x)$, $q(y)$ are probability distributions, $\Gamma(p,q)$ is the set of probabilistic distributions on $\mathbb{R} \times \mathbb{R}$ whose marginals represent $p$ and $q$ on the first and second moments, respectively.

\subsubsection{HEMS Models}
The previous techniques measured the quality of the generated data based on its distance from the probability distribution of the real data. In this technique, we measure the quality of the synthetic data by its ability to be used in the training of a Q-learning-based smart home energy management system (HEMS). We implemented the Q-learning model in the MATLAB platform. The ESS fixed charging/discharging power is 4 $kW$, and the ESS capacity is 16 $kW\cdot h$. The learning rate is 0.8, the discount factor is 0.7, and the initial $\epsilon$ value for the $\epsilon$-greedy exploration algorithm is 0.05. The energy trading price follows a fixed pattern as in \cite{z1}. We recorded the average results over 10 randomized runs.

\subsection{Distance Metrics Evaluation Results}
\label{results}
We used the KL-divergence, MMD, and Wasserstein distance metrics to measure how close is the distribution of synthetic data to the real data. Table~\ref{tab:metricresults} shows these statistical metrics for smart home electrical load consumption, PV production, and EV charging load consumption synthetic data generated by the GMM, GAN, and VAE-GAN generative models. For all metrics, the lower the errors are the better, corresponding to a closer match between the synthetic and real data. 

The results allow us to draw several conclusions. First, we find that\noteblue{all three models have relatively close errors for synthetic EV charging datasets, which can be explained by the comparative regularity of EV charging datasets versus PV generation, which is influenced by the weather or load, depending on many personal choices. Another possible explanation is that PV generation and load consumption training data were almost twice the size of EV charging data, suggesting that rich historical training data could have a positive impact on accuracy.} Second, we find that for the KL-divergence and Wasserstein distance, our VAE-GAN-based approach \noteblue{almost always provides} the best performance. For the MMD, in contrast, the best performance is provided by the GAN approach. This reversal in the relative order of the generator approaches for MMD is likely a result of the choice of mean embedding features. Understanding the exact mechanism of this difference requires future work.

\begin{table*}
\captionsetup{justification=centering}
\caption{Distance between real and synthetic smart grid data distribution using KL-divergence, Wasserstein distance, and MMD. Best results are highlighted in \textbf{bold}.}
\label{tab:metricresults}
\resizebox{\textwidth}{!}{%
\noindent\begin{tabularx}{\textwidth}{@{}l*{10}{C}c@{}}
\toprule
\multicolumn{1}{c}{\multirow{2}{*}{\textbf{Model}}} &
\multicolumn{3}{c}{KL divergence} & \multicolumn{3}{c}{Wasserstein distance} & \multicolumn{3}{c}{MMD} \\ \cmidrule(l){2-4} \cmidrule(l){5-7} \cmidrule(l){8-10}
\multicolumn{1}{c}{} & \multicolumn{1}{c}{Load} & \multicolumn{1}{c}{PV} & \multicolumn{1}{c}{EV} &
\multicolumn{1}{c}{Load} & \multicolumn{1}{c}{PV} & \multicolumn{1}{c}{EV} & 
\multicolumn{1}{c}{Load} & \multicolumn{1}{c}{PV} & \multicolumn{1}{c}{EV} \\ \midrule
GMM  & 0.277 & 1.067 & 0.127 & 619.3 & 1319.3 & 0.006  & 0.168 & 0.074 & 0.000047 \\
\addlinespace
GAN  & 0.104 & 0.454 & 0.104 & 437.42 & 1080.5 & \textbf{0.002} & \textbf{0.010} & \textbf{0.049} & \textbf{0.000028} \\
\addlinespace
VAE-GAN & \textbf{0.065} & \textbf{0.071} & \textbf{0.085} & \textbf{295.458} & \textbf{648.87} & 0.005 & 0.011 & 0.186 & 0.000071 \\
\bottomrule
\end{tabularx}%
}
\vspace{-2mm}
\end{table*}

\begin{figure*}
    \centering
    \captionsetup{justification=centering}
    \begin{subfigure}{0.30\textwidth}
        \includegraphics[width=\linewidth]{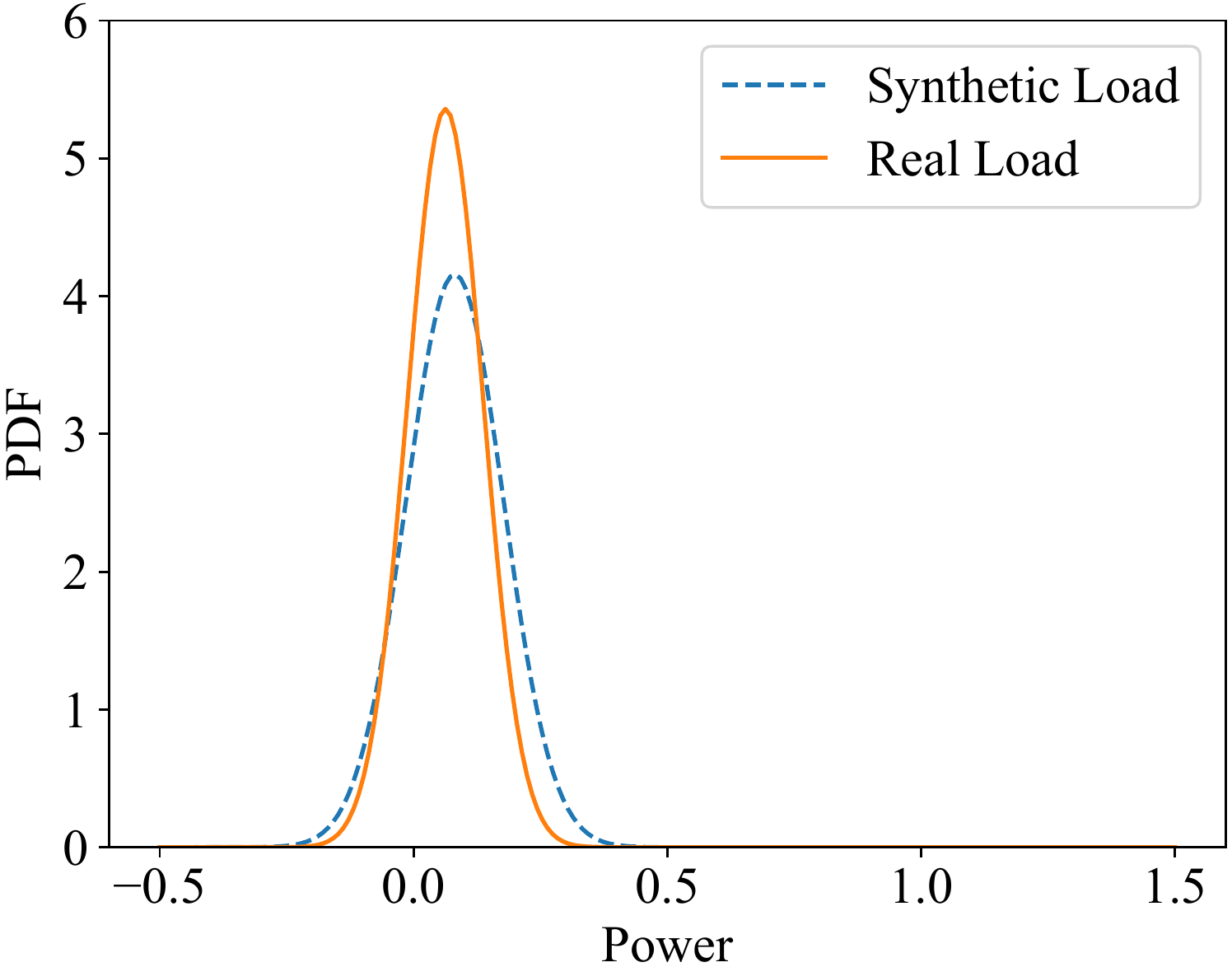}
        \subcaption{GMM}
        \label{fig:gmmloaddist}
    \end{subfigure}%
    \hfill%
    \begin{subfigure}{0.30\textwidth}
        \includegraphics[width=\linewidth]{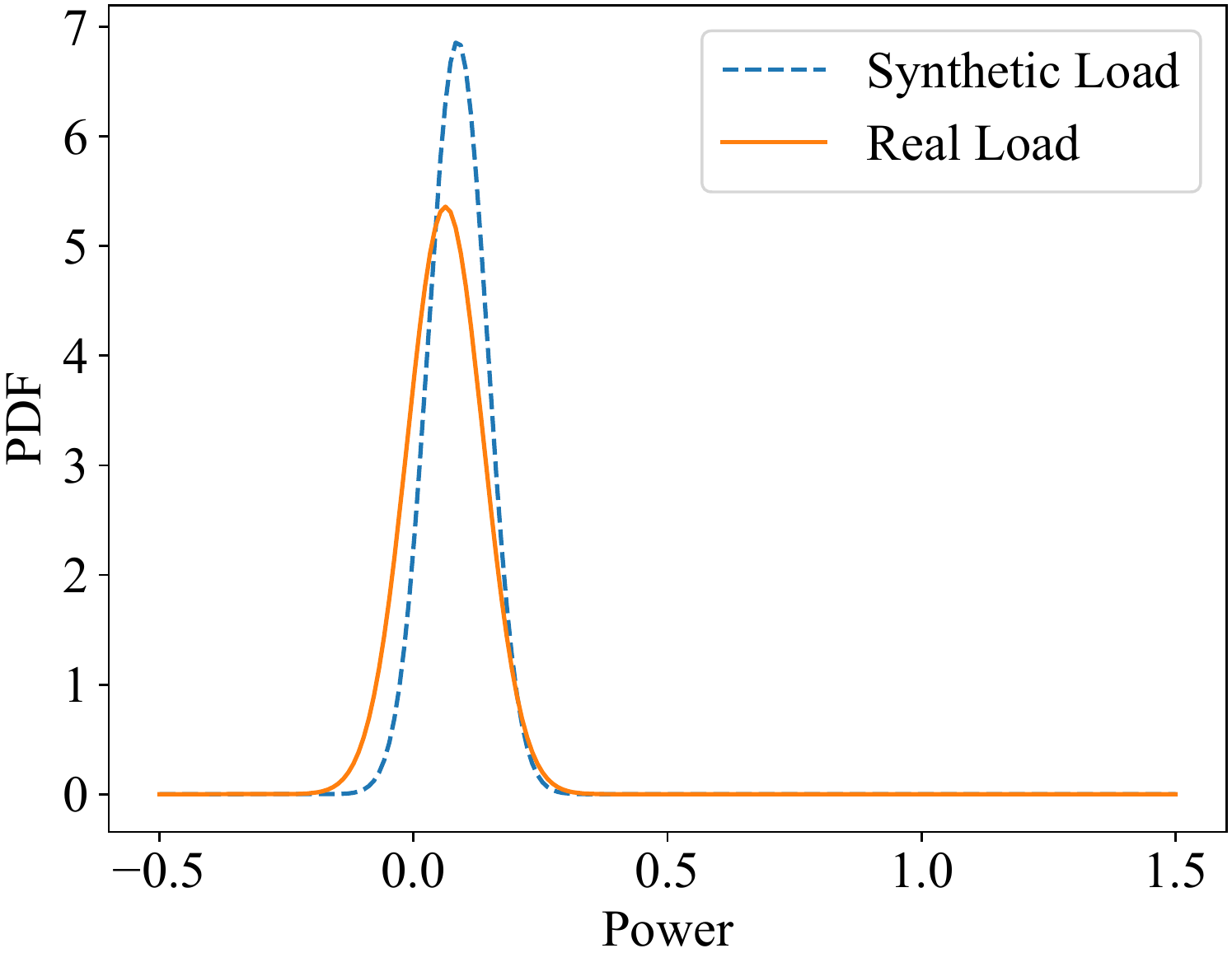}
        \subcaption{GAN}
        \label{fig:ganloaddist}
    \end{subfigure}%
    \hfill%
    \begin{subfigure}{0.30\textwidth}
        \includegraphics[width=\linewidth]{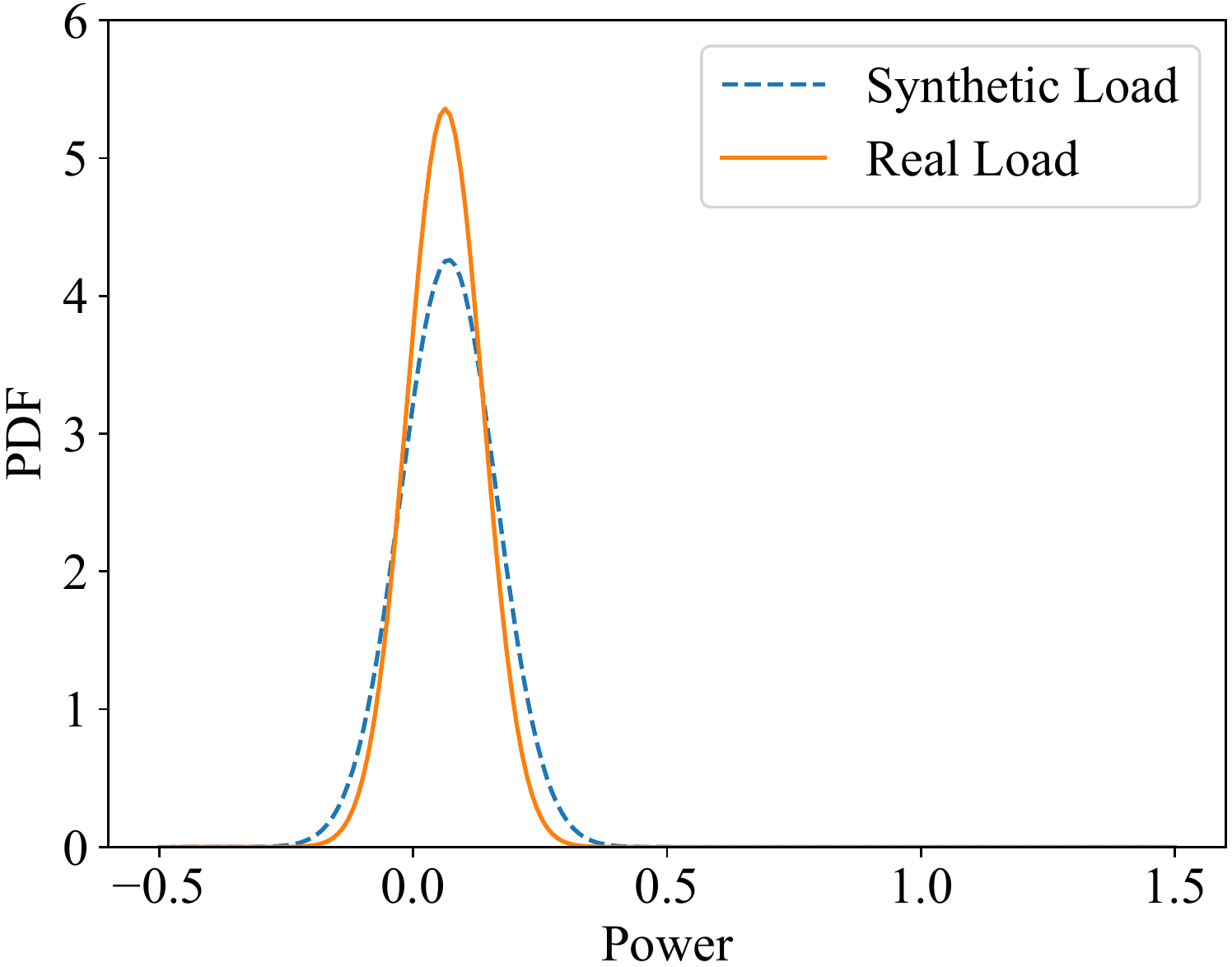}
        \subcaption{VAE-GAN}
        \label{fig:vaeganloaddist}
    \end{subfigure}%
    \caption[]{\small Electrical load consumption real and synthetic data probability density for (a) GMM, (b) GAN, and (c) VAE-GAN generative models. The orange line shows the real data PDF, the blue line shows the synthetic data PDF.}
    \label{fig:loaddists}
\end{figure*}

\begin{figure*}
    \centering
    \captionsetup{justification=centering}
    \begin{subfigure}{0.3\textwidth}
        \includegraphics[width=\linewidth]{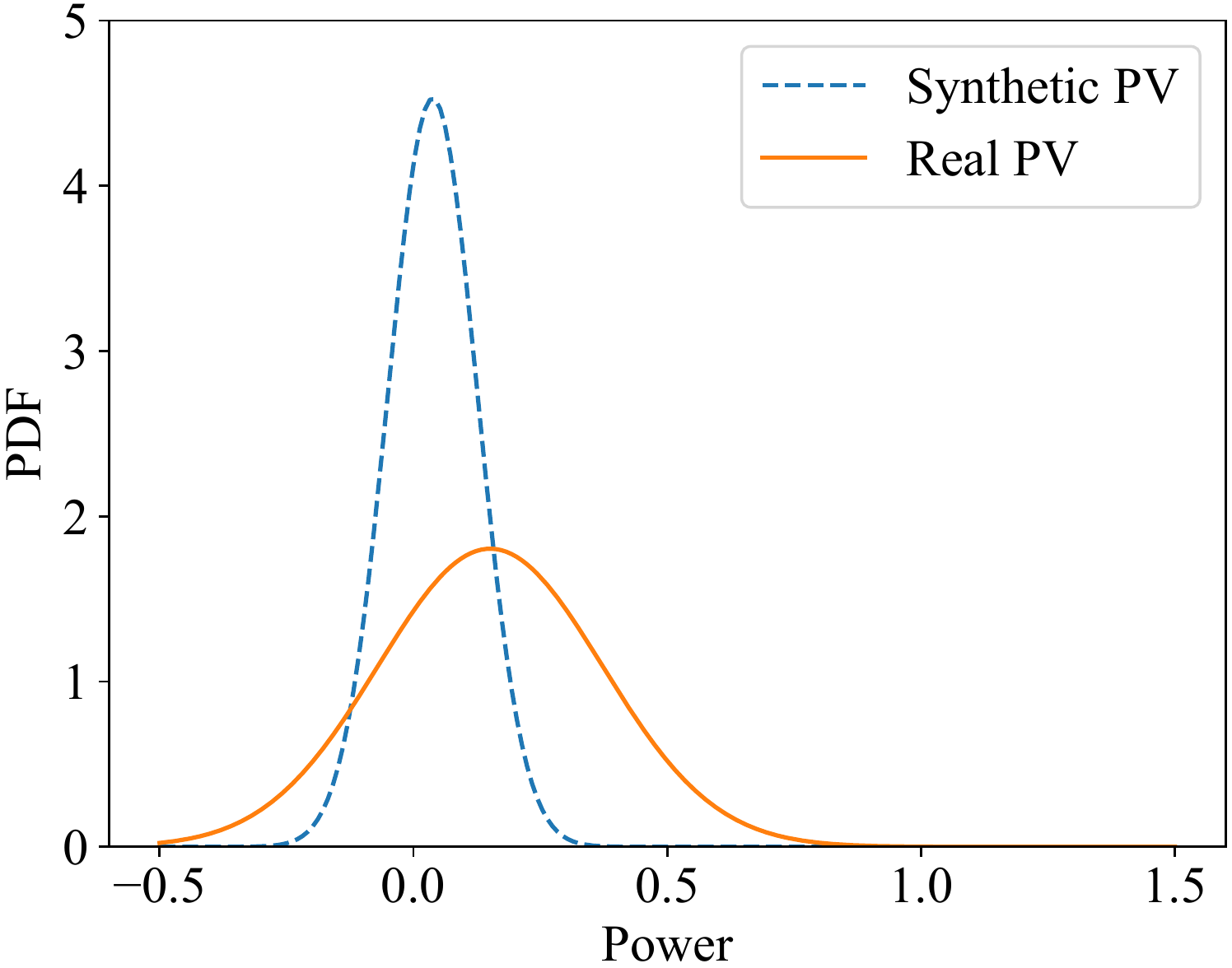}
        \subcaption{GMM}
        \label{fig:gmmpvdist}
    \end{subfigure}%
    \hfill%
    \begin{subfigure}{0.3\textwidth}
        \includegraphics[width=\linewidth]{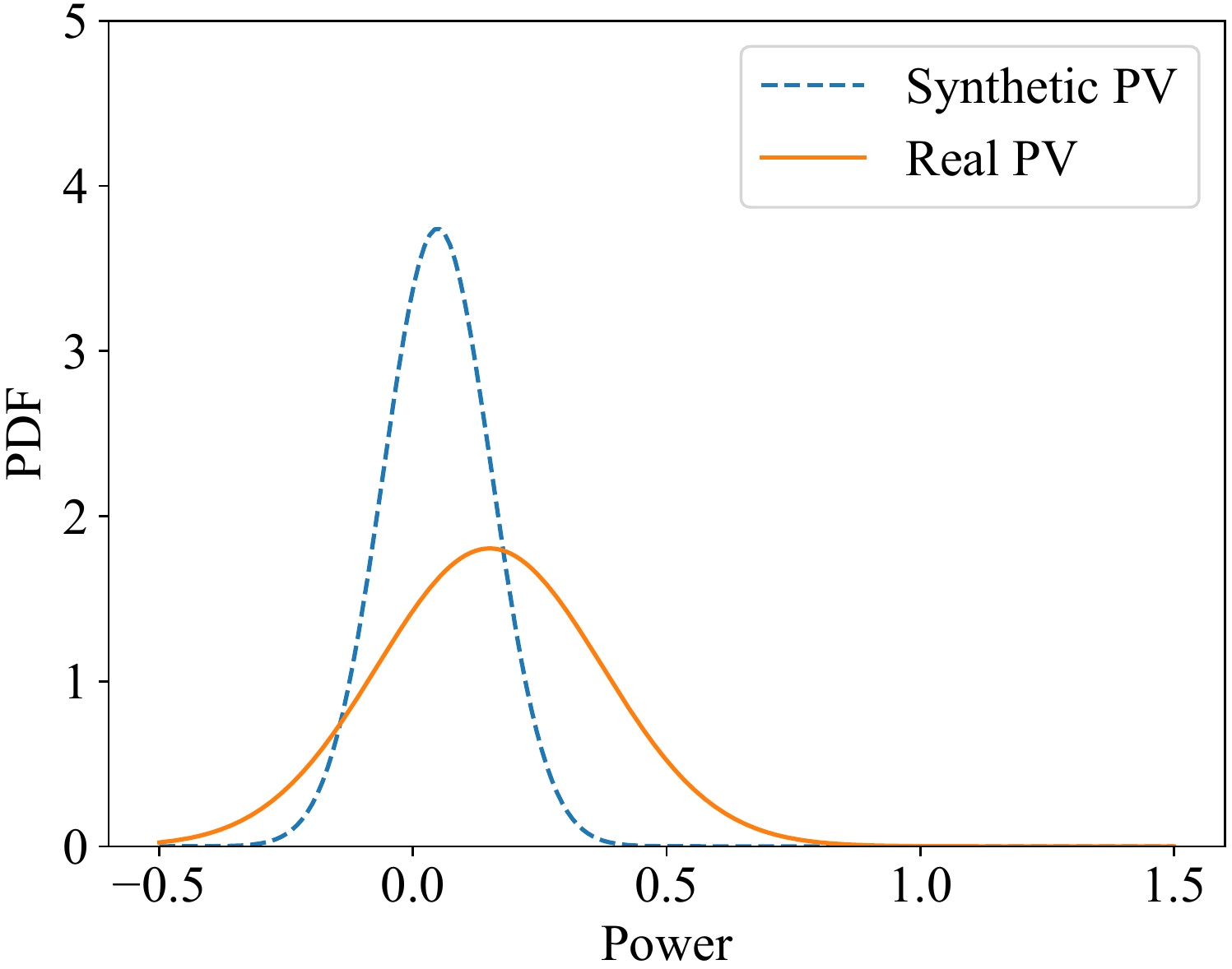}
        \subcaption{GAN}
        \label{fig:ganpvdist}
    \end{subfigure}%
    \hfill%
    \begin{subfigure}{0.3\textwidth}
        \includegraphics[width=\linewidth]{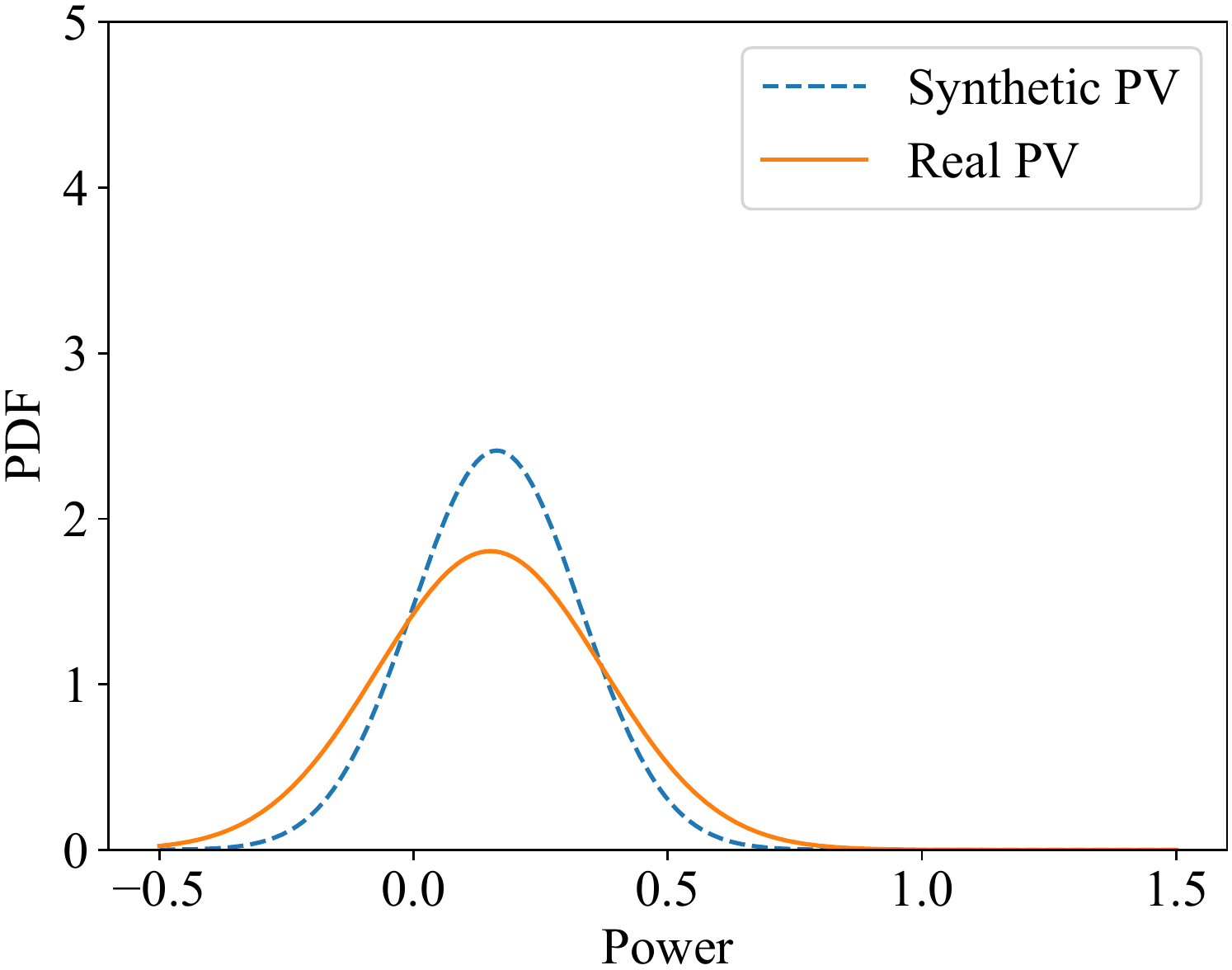}
        \subcaption{VAE-GAN}
        \label{fig:vaeganpvdist}
    \end{subfigure}%
    \caption[]{\small PV power production real and synthetic data probability density for (a) GMM, (b) GAN, and (c) VAE-GAN generative models. The orange line shows the real data PDF, the blue line shows the synthetic data PDF.}
    \label{fig:pvdists}
\end{figure*}

\begin{figure*}
    \centering
    \captionsetup{justification=centering}
    \begin{subfigure}{0.3\textwidth}
        \includegraphics[width=\linewidth]{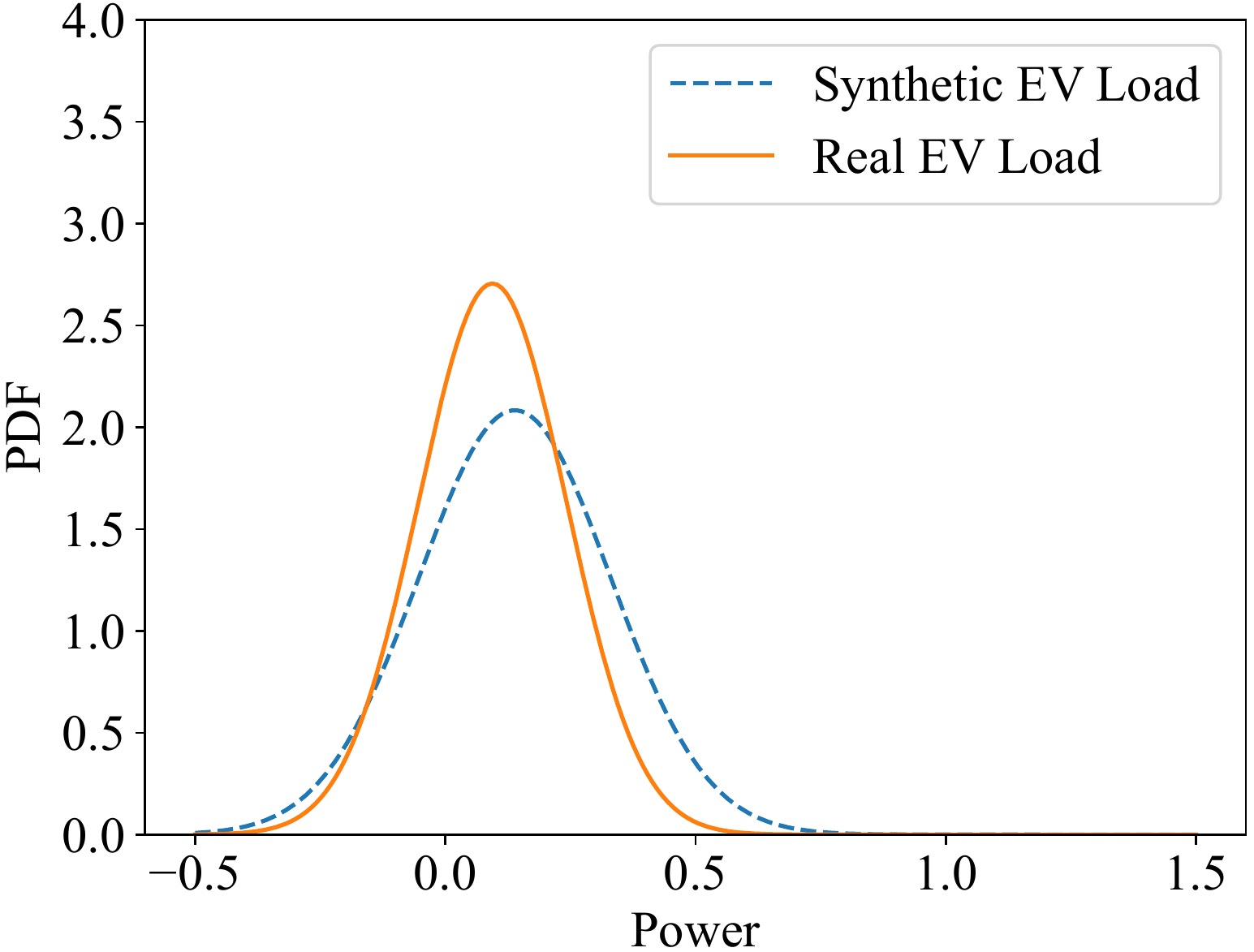}
        \subcaption{GMM}
        \label{fig:gmmevdist}
    \end{subfigure}%
    \hfill%
    \begin{subfigure}{0.3\textwidth}
        \includegraphics[width=\linewidth]{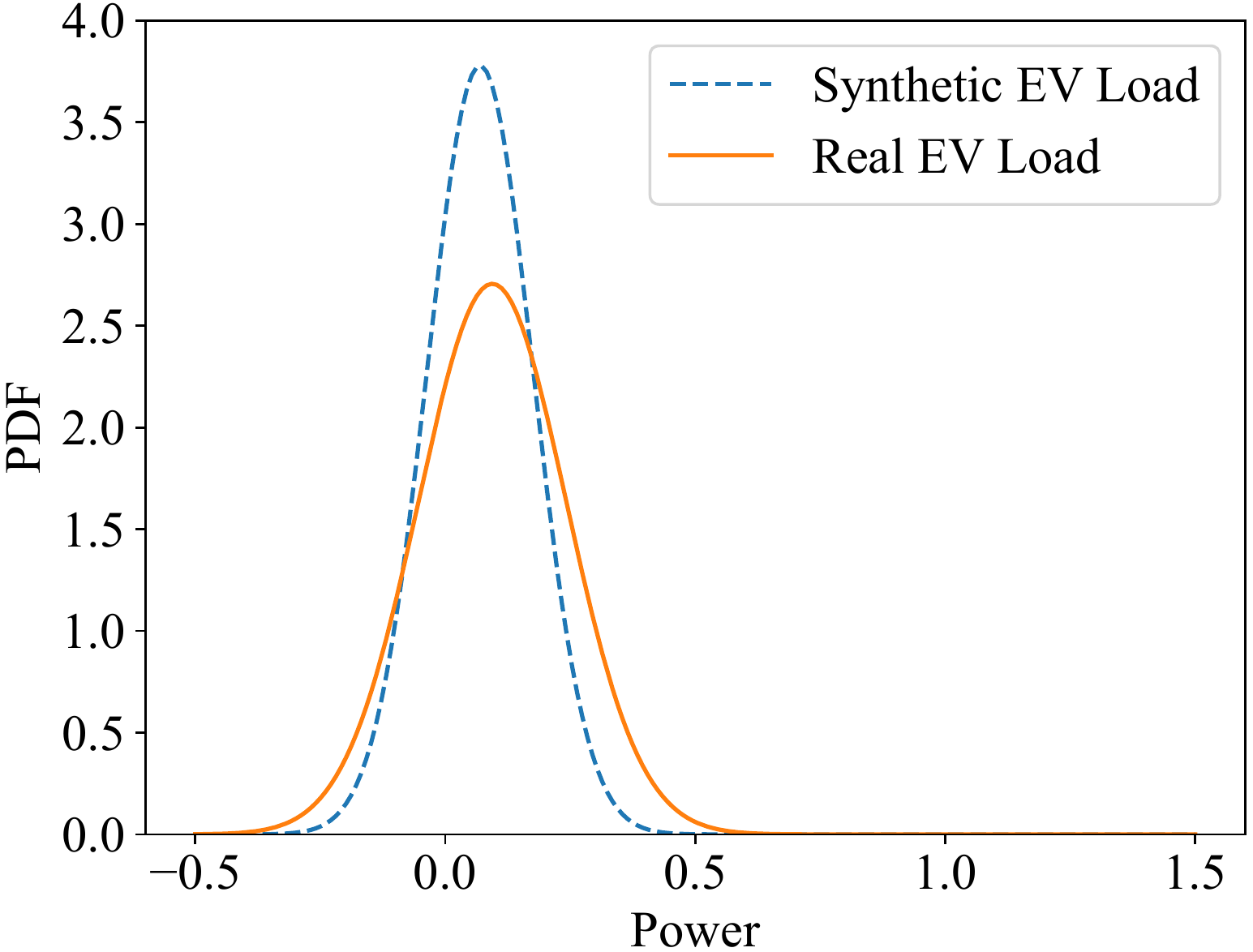}
        \subcaption{GAN}
        \label{fig:ganevdist}
    \end{subfigure}%
    \hfill%
    \begin{subfigure}{0.3\textwidth}
        \includegraphics[width=\linewidth]{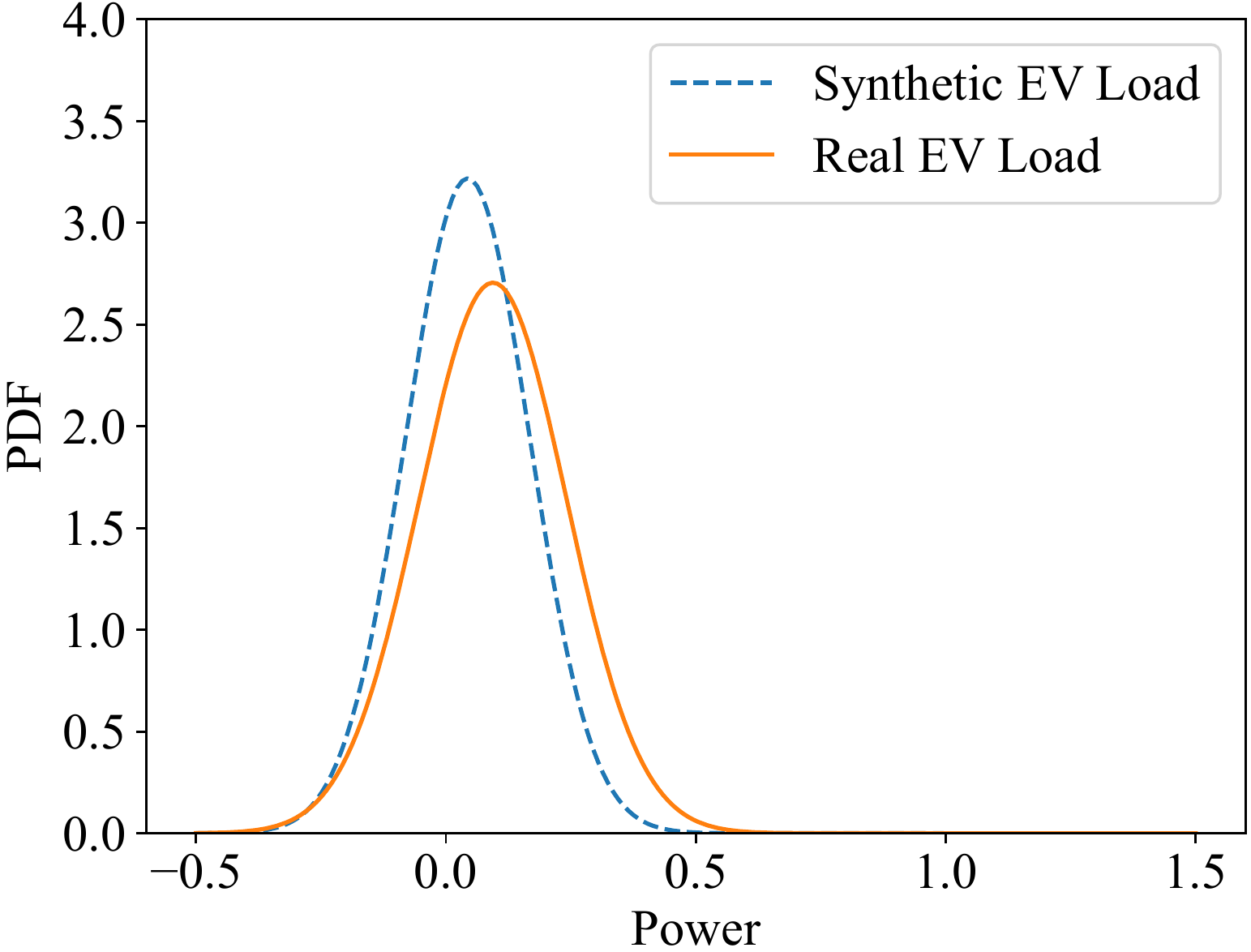}
        \subcaption{VAE-GAN}
        \label{fig:vaeganevdist}
    \end{subfigure}%
    \caption[]{\small EV charging load consumption real and synthetic data probability density function for (a) GMM, (b) GAN, and (c) VAE-GAN generative models. The orange line shows the real data PDF, the blue line shows the synthetic data PDF.}
    \label{fig:evdists}
    \vspace{-2mm}
\end{figure*}

While single numerical metrics quantifying the distance between the probability density functions are useful, examining the distributions as a whole provides us with a better understanding of the quality of the generated data. \Cref{fig:loaddists,fig:pvdists,fig:evdists} show the Probability Density Functions (PDF) for real and synthetic smart home data. 

\begin{figure*}
    \centering
    \vspace{-5mm}
    \begin{subfigure}{0.8\textwidth}
        \includegraphics[width=\linewidth]{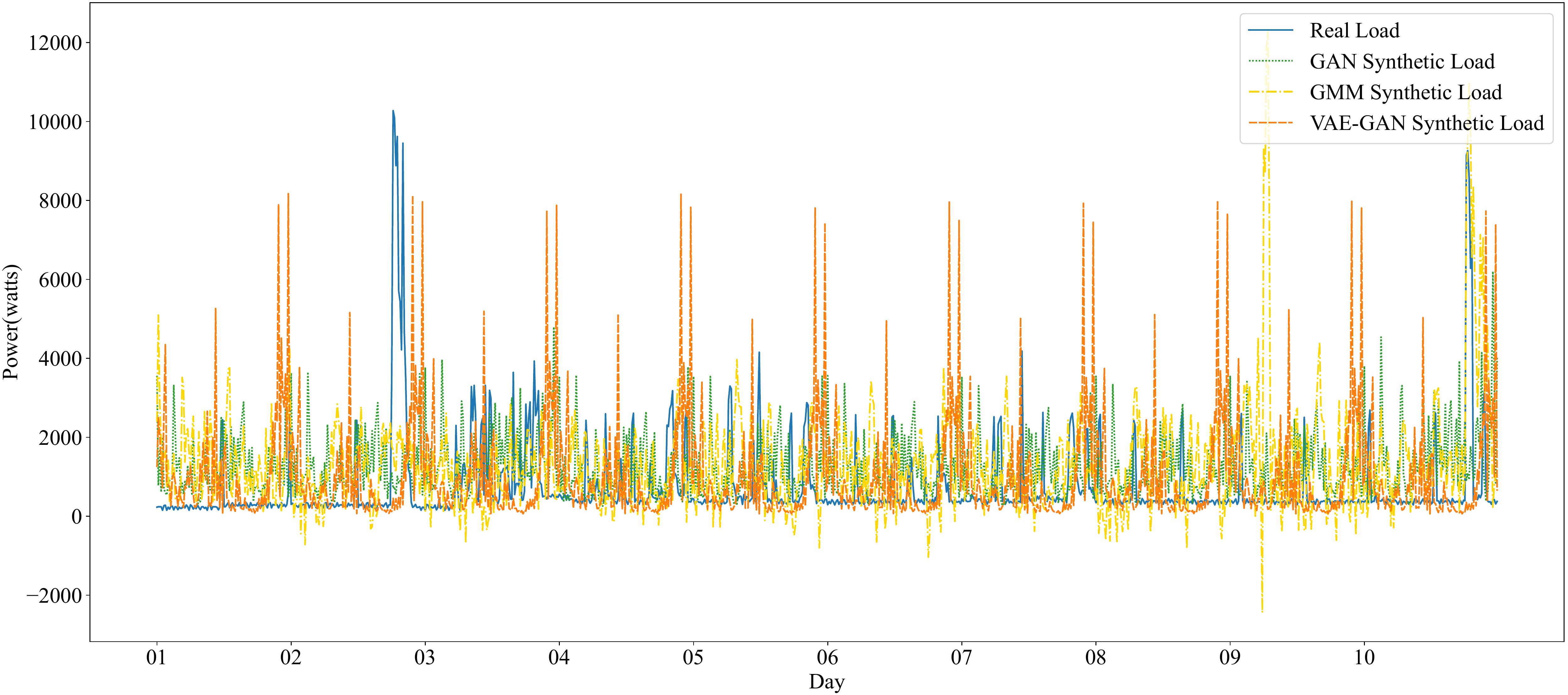}
        \subcaption{Aggregated Load}
        \label{fig:loadscatter}
    \end{subfigure}
    
    \begin{subfigure}{0.8\textwidth}
        \includegraphics[width=\linewidth]{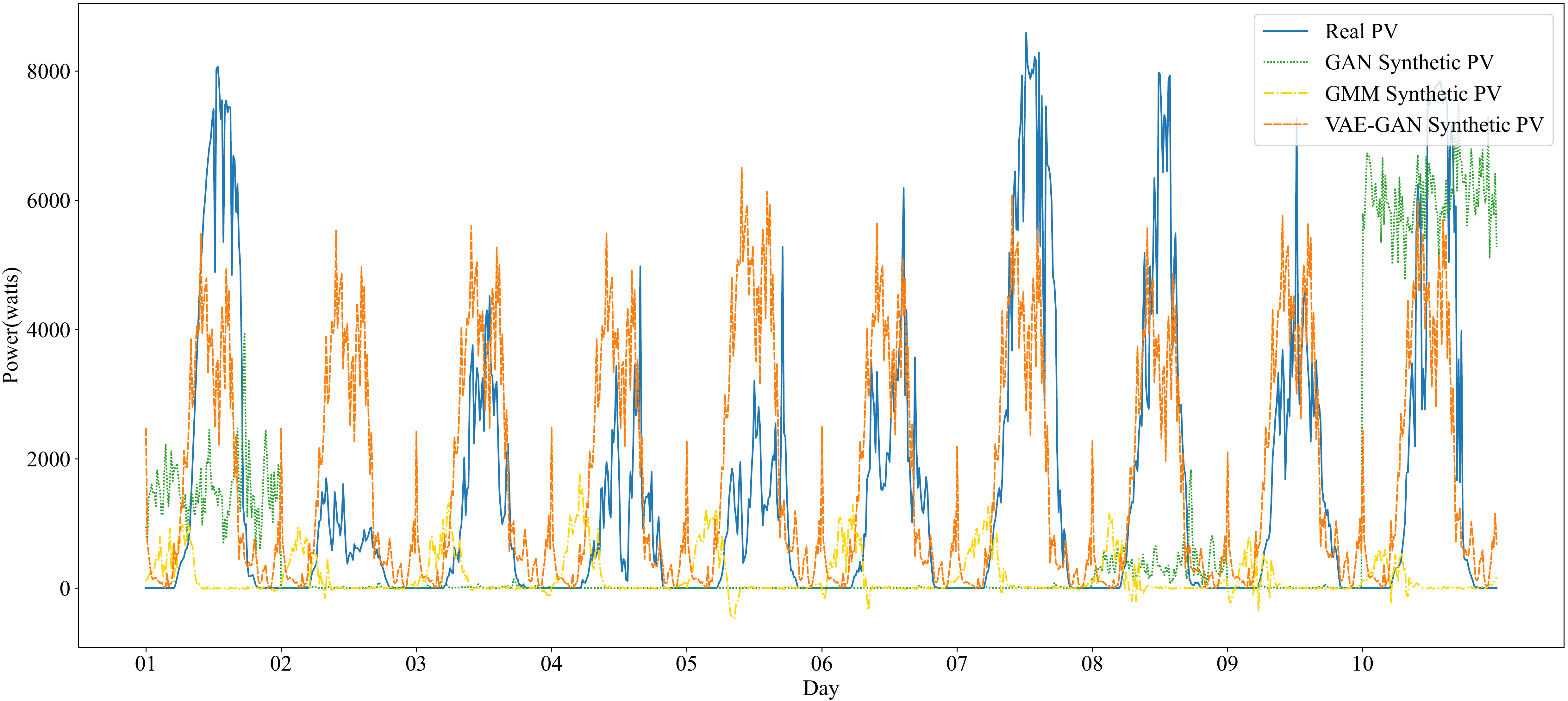}
        \subcaption{PV Power Production}
        \label{fig:pvscatter}
    \end{subfigure}
    
    \begin{subfigure}{0.8\textwidth}
        \includegraphics[width=\linewidth]{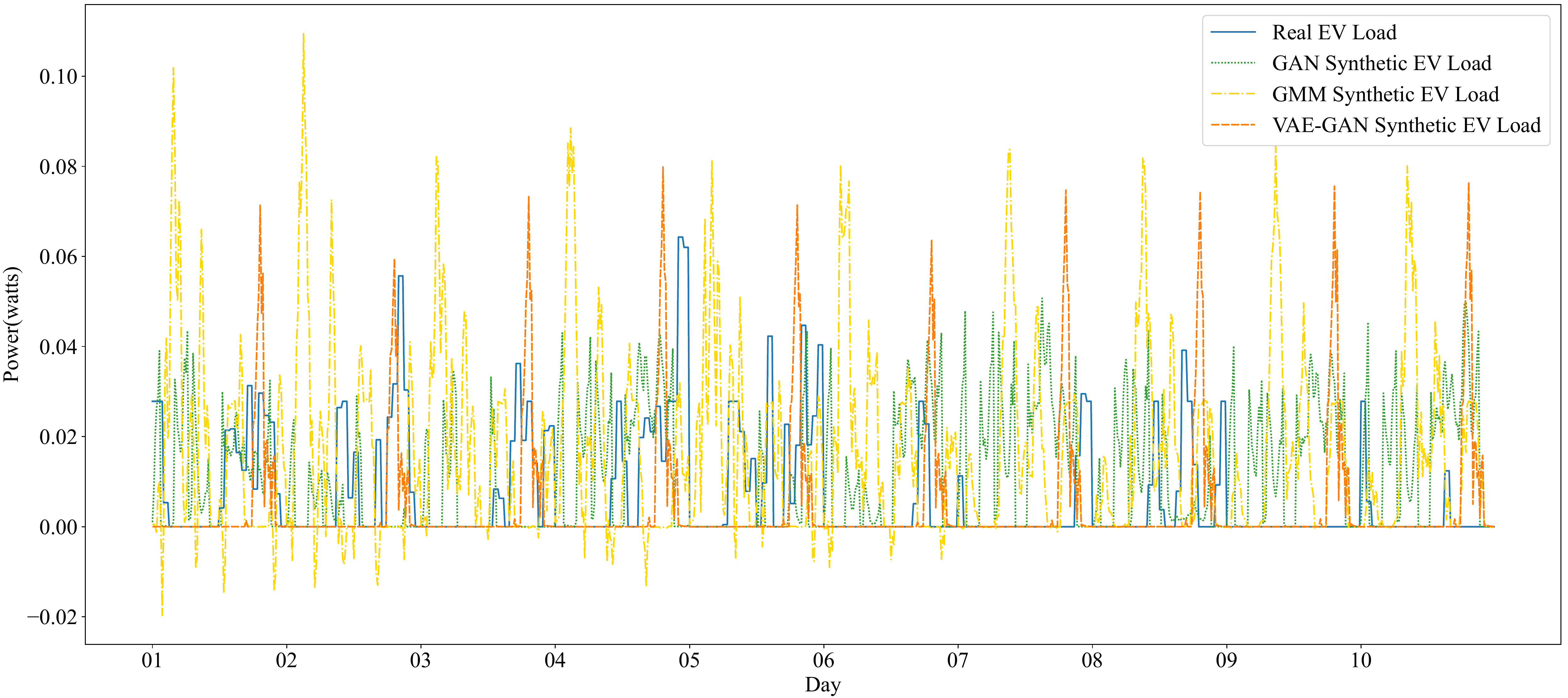}
        \subcaption{EV Load Consumption}
        \label{fig:evscatter}
    \end{subfigure}
    \caption[]{\small A sample of 10 days of synthetic data generated from GAN, GMM, and VAE-GAN models compared with the real test data }
    \label{fig:vaeganscatter}
\end{figure*}

Fig.~\ref{fig:loaddists} shows the PDF for normalized real and synthetic electrical load consumption for each generative model. While both the GMM and VAE-GAN distributions are relatively close to the real data, the synthetic data generated by VAE-GAN provides the best matches with the real data distribution in terms of the standard deviation from the mean value, as supported by the distance metrics. Fig.~\ref{fig:loadscatter} represents the pattern of real and synthetic electrical load consumption during the test data. It is evident from the close similarity of these patterns that the VAE-GAN network is capable of learning the distribution and patterns of smart home aggregated load consumption data and producing samples that reflect the same characteristics.

Fig.~\ref{fig:vaeganpvdist} illustrates that the VAE-GAN model performs better compared to two other generative models in generating synthetic PV production data, a conclusion that is supported by Fig.~\ref{fig:pvscatter}, where the pattern of synthetic PV production data generated by VAE-GAN is reasonably close to the pattern of real PV production data. We note that although PV production follows sunrise and sunset patterns, it is also highly affected by unpredictable environmental factors,\noteblue{therefore it is not expected for them to be the same equivalent.} 

Finally, Fig.~\ref{fig:evscatter} presents the EV charging load consumption real and synthetic data PDF for GMM, GAN, and VAE-GAN generative models. In Fig.~\ref{fig:gmmevdist} it is shown that synthetic EV charging load consumption generated by the GMM model is larger in power range compared to the real EV charging load consumption. On the other hand, the GAN model in Fig.~\ref{fig:ganevdist}, generates data centered around the average. Although the VAE-GAN model, Fig.~\ref{fig:vaeganevdist}, generates a slightly smaller power consumption than actual EV charging data, the distribution of load consumption is comparable to that of real data. Accordingly, each generative model shows the same characteristics in the sample EV data presented in Fig.~\ref{fig:evscatter}.

\subsection{HEMS Performance Evaluation Results}

\begin{figure*}
    \centering
        \centering
        \captionsetup{justification=centering}
        \includegraphics[width=0.9\textwidth]{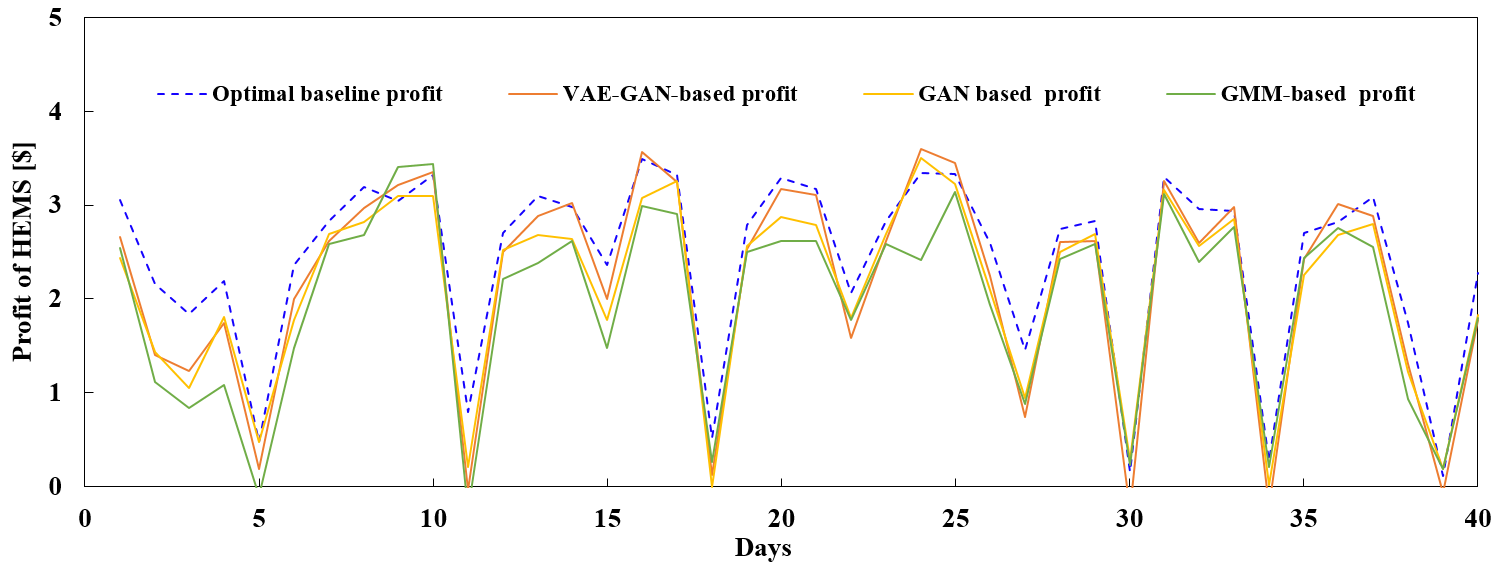}
    \caption[]{\small HEMS test performance comparison in 40 days by using real-world data (16 kWh ESS Capacity) \noteblue{The the total profit of optimal based, VAE-GAN, GAN, and GMM based methods are 97.54\$, 86.31\$, 85.18\$, and 79.36\$, respectively. }}
        \label{fig-pro}
\vspace{-2mm}
\end{figure*}

\begin{figure*}
    \centering
    \begin{subfigure}{0.45\textwidth}
        \includegraphics[width=\linewidth]{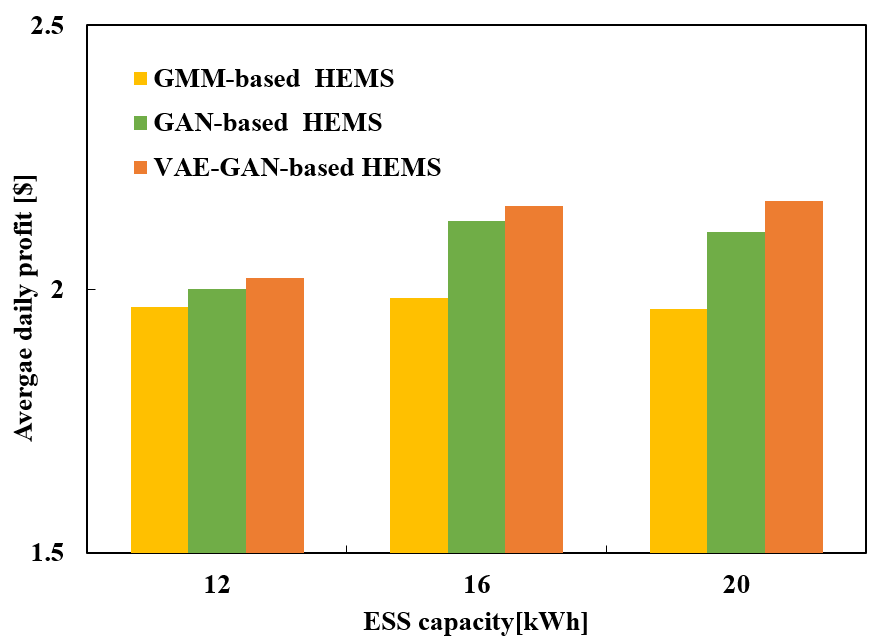}
        \subcaption{Average daily HEMS profit comparison under various ESS capacities}
        \label{fig-ess}
    \end{subfigure}%
    \qquad
    \begin{subfigure}{0.45\textwidth}
        \includegraphics[width=\linewidth]{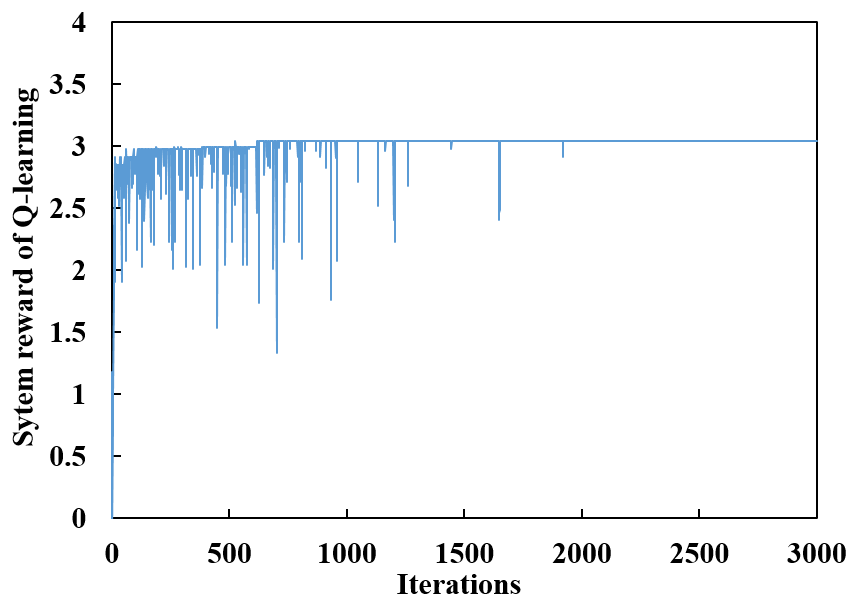}
        \subcaption{Convergence performance of Q-learning based HEMS in one episode}
        \label{fig-con}
    \end{subfigure}%
        \captionsetup{justification=centering}
        \caption[]{\small Smart home HEMS performance analyses}
    \label{fig:hemsanalysis}
\vspace{-3mm}
\end{figure*}

In this section, we investigate the performance of the Q-learning-based HEMS. For each type of synthetic data, we trained a corresponding policy, which\noteblue{was then} evaluated using real-world data (see Algorithm 1).\noteblue{We compare these policies against an optimal baseline that uses real-world data for both offline training and online operation testing, indicating that the Q-learning agent can best learn the hidden data patterns and prepare for the test.}\noteblue{The optimal baseline is expected to achieve the best overall performance by using the same real-world dataset for both training and testing, but it cannot guarantee that the optimal baseline can obtain higher profit every single day.}

Fig. \ref{fig-pro} shows the 40 days HEMS profit of different data generation methods. While the profit varies with the environmental conditions each day, as expected, the strategy trained on real-world data achieves the highest profit on most days. However, we found that the HEMS trained on the data generated by the proposed VAE-GAN model achieves a comparable performance. The GAN-based model achieves a somewhat lower performance, while the HEMS trained on the GMM-generated data shows the least profit. 

The performance of the HEMS also depends on the capacity of the energy storage system (ESS) of the smart home. The average profit of the differently trained HEMS systems for various ESS capacities is shown in Fig. \ref{fig-ess}. As expected, a larger ESS improves the profit of the HEMS, as the agent has higher flexibility in making decisions. We found that for all ESS values, the VAE-GAN-based HEMS performs best, followed by the GAN and GMM-based methods. The higher the ESS capacity increases the advantage of the VAE-GAN compared to the other methods. 


Finally, for practical applications, it is essential to establish that the learning process is stable. Fig.~\ref{fig-con} shows the reward of the Q-learning process function of the learning iterations. The figure shows that the agent has a stable convergence after the exploration phase.

\subsection{\noteblue{Complexity and Feasibility Analysis}}
\noteblue{Based on the simulation results, this section will discuss the complexity and feasibility of the proposed scheme, including a) feasibility and scalability, b) implementation complexity, c) management costs, and d) economic benefits. 
\begin{itemize}
    \item Feasibility and scalability: Compared with conventional model-based methods such as Markov chain and simulators, the applied data-driven method has higher scalability and can be easily generalized to other scenarios. In the experiment, note that the only requirement for applying the proposed methods is the dataset, indicating that it can be easily generalized from smart homes to smart builds and communities with proper datasets. By contrast, it is impractical to build very huge simulators or Markov chains to describe all the detailed behaviors of a large area.
    \item Organizational complexity: The main complexity of the proposed method is the neural network architecture and design, which is a well-known issue for ML-based methods. However, in this work, note that no dedicated models are designed for any household devices or energy consumption behaviors, significantly reducing the organizational complexity.       
    \item Management costs: As above-analyzed, the dataset is the only requirement of the proposed method. When the physical system changes, e.g., adding or removing new devices, our model can be easily updated by feeding new datasets to the neural networks, which means a very low system management cost.  
    \item Economic benefits: The simulation results in Fig.\ref{fig-pro} and \ref{fig-ess} have demonstrated that the proposed method can obtain satisfying profit for the smart home operation. The benefit can be greater when considering larger application scenarios such as smart buildings and communities.
\end{itemize}
} 
\section{Conclusion}
\label{Conclusion}

\noteblue{In recent years, smart home management increasingly takes advantage of advanced 
data-demanding machine learning approaches, but the availability of find-grained data sets may prevent the applications.  
Therefore, synthetic data generation has become a critical enabler for smart home management. 
In this paper, we present a variational autoencoder GAN (VAE-GAN) approach for the synthetic data generation of smart home datasets. We have shown that the proposed VAE-GAN method can generate high-quality data that better matches the statistical properties of real-world data compared to benchmarks. 
In addition, we use the synthetic data to train a Q-learning-based smart home energy management system (HEMS), achieving a higher profit than other data-generation methods.  
The simulations prove that the generated synthetic data can be used for HEMS offline training, and the agent obtains a satisfying real-world online performance without touching the real-world data.}
\section*{Acknowledgement}
Hao Zhou and Melike Erol-Kantarci were supported by the Natural Sciences and Engineering Research Council of Canada (NSERC), Collaborative Research and Training Experience Program (CREATE) under Grant 497981 and Canada Research Chairs Program.

\bibliographystyle{IEEEtran}
\bibliography{7_Bibliography.bib}
\end{sloppypar}
\end{document}